\documentclass[acmlarge]{acmart}
\AtBeginDocument{%
  }
\usepackage{multirow}
\usepackage{float}
\usepackage{graphicx}
\usepackage{subcaption}

\setcopyright{acmlicensed}
\copyrightyear{2025}
\acmYear{2025}
\acmDOI{XXXXXXX.XXXXXXX}

\acmJournal{POMACS}
\acmVolume{37}
\acmNumber{4}
\acmArticle{111}
\acmMonth{8}

\begin{document}

\title{Prediction of Challenging Behaviors Associated with Profound Autism in a Classroom Setting Using Wearable Sensors}

\author{Yadhu Kartha}
\email{yadhukartha@gatech.edu}
\orcid{0000-0003-2914-8087}
\authornotemark[1]
\affiliation{%
  \institution{Georgia Institute of Technology}
  \city{Atlanta}
  \state{Georgia}
  \country{USA}}

\author{Conor Anderson}
\affiliation{%
  \institution{The Center For Discovery}
  \city{Harris}
  \state{New York}
  \country{USA}}

\author{Jenny Foster}
\affiliation{%
 \institution{The Center For Discovery}
  \city{New York City}
  \state{New York}
  \country{USA}}

\author{Theresa Hamlin}
\affiliation{%
  \institution{The Center For Discovery}
  \city{Harris}
  \state{New York}
  \country{USA}}

\author{Johanna Lantz}
\affiliation{%
  \institution{The Center For Discovery}
  \city{Harris}
  \state{New York}
  \country{USA}}

\author{Ryan Lay}
\affiliation{%
  \institution{The Center For Discovery}
  \city{Harris}
  \state{New York}
  \country{USA}}

\author{Juergen Hahn}
\affiliation{%
  \institution{Rensselaer Polytechnic Institute}
  \city{Troy}
  \state{New York}
  \country{USA}}
\email{hahnj@rpi.edu}
\orcid{0000-0002-1078-4203}

\author{Gari D. Clifford}
\affiliation{%
  \institution{Georgia Institute of Technology, Emory University}
  \city{Atlanta}
  \state{Georgia}
  \country{USA}}

\author{Hyeokhyen Kwon}
\affiliation{%
  \institution{Georgia Institute of Technology, Emory University}
  \city{Atlanta}
  \state{Georgia}
  \country{USA}}
\email{hyeokhyen.kwon@emory.edu}
\orcid{0000-0002-5693-3278}

\renewcommand{\shortauthors}{Kartha et al.}

\begin{abstract}
Autism Spectrum Disorder (ASD) is characterized by challenges with social interaction and communication and by restricted or repetitive patterns of thought and behavior, with significant variability in presentation.
Approximately a quarter of children with ASD are classified as having profound autism, who often exhibit challenging behaviors, such as self-injurious behavior, aggression, elopement, or pica, that pose serious safety risks and disrupt learning in educational settings.
Prior work has applied wearable sensors and machine learning to detect challenging behaviors, but has been largely confined to controlled laboratory environments.
This work demonstrates that predicting challenging behavior episodes is feasible in a real-world special education classroom.
We collected approximately 110.7 hours of labeled multimodal wearable data comprising accelerometry, electrodermal activity (EDA), and skin temperature from 9 children and young adults aged 10 to 21 years across standard classroom sessions.
We fine-tuned state-of-the-art foundation models for multimodal wearable time-series analysis and show that challenging behavior episodes can be predicted up to 10 minutes in advance with an AUC-ROC of 0.78.
These results establish a concrete foundation for developing proactive in-class intervention systems that enable teachers to minimize the safety risks of challenging behaviors in special education classrooms.
\end{abstract}



\maketitle

\section{Introduction}
Autism spectrum disorder (ASD) is a neurodevelopmental condition characterized by challenges in social communication and interactions and the presence of restricted or repetitive behaviors \cite{hodges_autism_2020}.
Among individuals with ASD, approximately one quarter is estimated to have profound autism, which is a subgroup that is especially vulnerable, often exhibiting frequent and intense episodes of self-injurious behavior, aggression, and agitation.
These challenging behaviors not only disrupt learning but also present serious safety concerns for the individuals themselves and for caregivers, peers, and educators.
Even when proactive strategies are in place, staff may need to use reactive strategies including physical restraint or isolation in order to prevent injury or property damage.
Observation is essential for identifying environmental influences on behavior and evaluating treatment outcomes. 
These observations are typically conducted in person or via video review. However, they are time-intensive and susceptible to subjective bias, often resulting in low interrater reliability \cite{rad_motion_2025}. 
Moreover, physiological responses to environment stimuli that often precede challenging behavior are not able to be observed using these traditional methods, yet such information can provide valuable insights for guiding  assessment and treatment.
As the need grows for scalable, objective, and context-aware monitoring tools, wearable sensing technologies have gained increasing attention.
These systems enable continuous, non-invasive recording of multimodal physiological and behavioral signals—such as accelerometry, electrodermal activity (EDA), and skin temperature, directly in real-world environments, offering a powerful alternative to conventional methods \cite{francese_supporting_2022,gao_wearable_2024,hernandez-capistran_commercial_2024}.
By facilitating fine-grained, temporally resolved monitoring, wearables have the potential to transform behavior tracking in profound autism from episodic documentation to real-time, data-driven intervention.

Human activity recognition (HAR) using wearable sensor data has seen significant advances over the past decade, evolving from early-stage methods based on hand-engineered features and classical machine learning (ML) algorithms such as decision trees and support vector machines \cite{plotz_automatic_2012}, to modern deep learning approaches that learn hierarchical and temporal patterns directly from raw sensor signals \cite{zhang_deep_2022}.
This shift has enabled the modeling of complex multimodal time-series data from sources such as accelerometers, electrodermal activity, and skin temperature, with improved robustness across environments and populations.
In particular, transformer-based models have emerged as a powerful alternative to convolutional and recurrent neural networks, leveraging self-attention mechanisms to capture long-range dependencies and achieve state-of-the-art performance in diverse HAR benchmarks \cite{dirgova_luptakova_wearable_2022}.
Complementing these architectural innovations, self-supervised learning (SSL) has addressed one of the field’s core limitations — the scarcity of labeled data — by enabling the extraction of meaningful representations from massive unlabeled datasets.
Pretraining on large-scale corpora, such as hundreds of thousands of person-days of wearable data, has led to substantial performance gains in downstream tasks with limited supervision, particularly in health and behavioral domains \cite{yuan_self-supervised_2024}.
However, deployment in ASD-related applications remains nascent.
Most existing work continues to focus on post hoc behavior detection under controlled conditions, with limited generalizability to naturalistic settings and almost no capacity for proactive prediction \cite{zwilling2022prediction}.
This gap is particularly acute for challenging behaviors in profound autism, where behavioral heterogeneity and noisy physiological signals present additional modeling challenges.

Despite promising advances in wearable-based behavior detection for autism, most existing approaches remain limited by single-modality data, shallow architectures, and retrospective classification frameworks that fail to anticipate behaviors in advance \cite{shamhan_advancements_2025,francese_supporting_2022}.
Prior studies have primarily leveraged accelerometry or EDA in isolation for detection, often neglecting the complementary information embedded in physiological signals such as skin temperature, which can modulate electrodermal responsiveness and improve affective state inference \cite{VANDOOREN2012298}.
Moreover, while transformer-based models have demonstrated success in human activity recognition \cite{dirgova_luptakova_wearable_2022}, their application to behavior prediction — particularly in clinical, low-resource settings — has been underexplored.
To overcome these limitations, we introduce a multimodal fusion framework that integrates accelerometry, EDA, and skin temperature signals using pretrained models fine-tuned for the \textit{prediction of challenging behaviors}.

\section{Related Works and Background}

\subsection{Understanding Behaviors in ASD}

Artificial intelligence (AI) has been extensively applied to autism spectrum disorder research across a range of sensing modalities.
In computer vision, convolutional and graph-based models have been used to analyze stereotypical movements and gesture patterns from video \cite{vyas_recognition_2019,siddiqui_wearable-sensors-based_2021}, while gaze and facial features have supported early ASD screening via deep learning pipelines \cite{lopez-martinez_wearable_2024}.
Speech and language analysis approaches extract acoustic and prosodic features using recurrent neural networks and spectrogram-based CNNs to distinguish ASD from neurotypical (NT) groups \cite{shamhan_advancements_2025}.
In parallel, wearable-based systems have gained traction for in-situation behavioral and physiological monitoring, with accelerometry and electrodermal activity (EDA) as the most commonly used modalities \cite{zhu_stress_2023}.
Accelerometer-based models range from traditional feature-based classifiers to deep sequential models such as LSTMs and temporal convolutional networks (TCNs), often applied to detect motor patterns or stress episodes \cite{zhu_stress_2023,rad_motion_2025}.
EDA, a proxy for sympathetic arousal, has been used in both ASD and general populations for stress classification using support vector machines (SVMs), random forests, and deep learning \cite{sanchez-reolid_machine_2022,zhu_feasibility_2022,yu_semi-supervised_2023}.
Some studies have incorporated skin temperature or heart rate variability, but typically within shallow fusion frameworks with limited modeling of temporal interactions \cite{logacjov_machine_2024,le_tran_thuan_machine_2024}.
Other approaches, albeit based upon data not collected from wearable devices, use information about sleep, Gastrointestinal issues, and prior behaviors to make long-term predictions about challenging behavior in profound ASD \cite{ferina_approaches_2025,ferina_predicting_2023}. However, this type of data and the prediction horizons are very different from the approach presented here. 
In contrast, our approach adapts deep temporal modeling and multimodal fusion complemented by post-hoc interpretability analyses to better capture behaviorally relevant signals over time, offering a framework for challenging behavior prediction.

\subsection{Sensor Representation Models}

\subsubsection{Accelerometer}
Self-supervised learning (SSL) has substantially advanced wearable time-series modeling by enabling representation learning from large-scale unlabeled datasets, reducing reliance on costly annotations.
Early SSL frameworks for wearable human activity recognition use objectives such as masked signal reconstruction \cite{yuan_self-supervised_2024}, contrastive learning \cite{ruan_ai_2025} across temporal segments, and sequence-level prediction \cite{zhang_deep_2022} to capture long-range temporal dependencies.
A notable example is pretraining on over 700,000 person-days of accelerometry data from the UK Biobank \cite{Sudlow2015UKBiobank}, where SSL models demonstrated strong cross-dataset generalization and consistent improvements on downstream activity recognition tasks, particularly under scarcity of labeled samples \cite{yuan_self-supervised_2024}.
Building on this paradigm, foundation models for wearable data have emerged, including the Pretrained Actigraphy Transformer (PAT), which adapts transformer architectures with patch-based embeddings for raw accelerometer signals \cite{ruan_ai_2025}, and accelerometer encoders distilled from high-fidelity photoplethysmography (PPG) representations using cross-modal alignment \cite{abbaspourazad_wearable_2025}.
These models established general-purpose encoders for health-related time-series data and achieved state-of-the-art performance in domains such as mental health assessment, sleep analysis, and cardiovascular monitoring \cite{abbaspourazad_wearable_2025}.
Accelerometry remains the most widely used modality in wearable HAR due to its low cost, ubiquity, and robustness, with earlier methods relying on handcrafted time- and frequency-domain features combined with classical classifiers such as support vector machines and random forests \cite{plotz_automatic_2012}.
In ASD research, accelerometer-based approaches have been used to detect stereotypical motor behaviors and challenging events, including SVM-based recognition of repetitive movements \cite{vyas_recognition_2019} and LSTM-based detection of self-injury and aggression in naturalistic classroom environments \cite{rad_motion_2025}.
However, most prior work remains focused on retrospective detection rather than forward prediction, often operates on small and highly individualized datasets, and does not leverage large-scale pretrained representations, limiting generalization across participants, settings, and behavioral contexts.
Our work addresses these limitations by adapting pretrained foundation models for accelerometry within a multimodal predicting framework, enabling anticipatory prediction of challenging behaviors in profound autism using limited labeled data collected in real-world educational settings.

\subsubsection{Electrodermal Activity}

Electrodermal activity is a widely used physiological marker of sympathetic nervous system activation, known for its sensitivity to emotional arousal, cognitive workload, and stress \cite{boucsein_electrodermal_2012}.
In wearable sensing, EDA is typically collected from the wrist or palm using surface electrodes, and it is commonly decomposed into tonic (slow-varying baseline) and phasic (event-related) components via techniques such as continuous decomposition analysis or convex optimization \cite{boucsein_electrodermal_2012}.
These components are used to derive hand-engineered features such as skin conductance level (SCL), number of peaks, response latency, and recovery slope, which are then fed into classical models including support vector machines, decision trees, and shallow neural networks.
Deep learning approaches have also been applied to stress prediction using EDA.
For example, Yu and Sano proposed a semi-supervised framework that combines labeled and unlabeled wearable data through consistency-based regularization to improve generalization in real-world, noisy environments \cite{yu_semi-supervised_2023}.
Their work demonstrated that EDA, when modeled with modern learning architectures and trained with both supervised and semi-supervised objectives, can yield robust stress prediction even under sparse labeling and heterogeneous conditions—challenges that are also central in the autism domain.
In the context of autism spectrum disorder, EDA has been explored for tracking arousal and stress-related behaviors, particularly in lab-based or short-duration studies.
Welch \textit{et al.} conducted a systematic review of 32 EDA studies in autism, identifying major limitations such as inconsistent preprocessing pipelines, non-standardized sensor placements, and lack of reproducibility across studies and devices \cite{welch_use_2022}.
Similarly, Sánchez-Reolid \textit{et al.} demonstrated the feasibility of EDA-based stress classification in non-ASD contexts using feature-based classical models, though their approach lacked resilience in dynamic, real-world environments \cite{sanchez-reolid_machine_2022}.
These limitations, i.e., reliance on hand-engineered features, lab-controlled conditions, and inconsistent preprocessing motivate our use of pretrained EDA encoders fine-tuned on naturalistic, in-situ data within a multimodal fusion framework.

\subsubsection{Skin Temperature}

Skin temperature has gained attention as a complementary physiological signal in wearable sensing due to its relationship with autonomic nervous system activity \cite{shin2025skintemp}.
Unlike faster-responding indicators such as electrodermal activity, skin temperature changes more gradually and reflects both environmental exposure and internal physiological shifts, such as vasoconstriction during stress \cite{shin2025skintemp}.
Typically sampled at low frequencies, it is often used as a contextual or auxiliary input rather than a primary driver of state classification.
Prior studies have leveraged temperature features in tandem with motion data to enhance behavioral state inference.
For example, Logacjov \textit{et al.} used accelerometer and skin temperature signals to classify sleep–wake cycles via random forests, showing that thermal signals improved classification robustness, particularly during low-movement periods \cite{logacjov_machine_2024}.
Le Tran Thuan \textit{et al.} incorporated skin temperature into stress classification pipelines using ensemble models (e.g., XGBoost and SVM), observing modest accuracy gains when temperature was fused with EDA and heart rate variability features \cite{le_tran_thuan_machine_2024}.
Despite this, temperature is often treated as a static or averaged covariate, with minimal modeling of its temporal dynamics or integration into deep learning architectures.
Additionally, its utility in neurodevelopmental contexts such as autism remains underexplored, partly due to concerns about signal reliability and the confounding effects of ambient temperature.
Our work advances this line by modeling skin temperature as a temporal signal within a multimodal fusion framework rather than a static covariate, enabling the capture of thermally structured patterns that may precede challenging behaviors.

\subsection{Multimodal Analysis}

Multimodal sensing, which integrates accelerometry, electrodermal activity, skin temperature, and other physiological signals, has emerged as a promising direction for behavior recognition and affective state modeling, particularly in health and mental health domains \cite{francese_supporting_2022,rad_motion_2025}.
By capturing complementary dimensions of human physiology — movement, arousal, and thermoregulation — multimodal systems offer the potential for improved robustness and interpretability compared to unimodal approaches.
Early fusion methods typically concatenate features from different modalities before input to a classifier, while late fusion strategies aggregate predictions from unimodal models \cite{Cui2025evaluating}.
However, many studies rely on shallow fusion techniques that use simple integration strategies, combining features or outputs from different modalities without deep joint modeling, thereby failing to capture modality-specific temporal dynamics or cross-modal interactions.
Our work addresses this limitation by systematically comparing fusion strategies of increasing complexity, from MLP-based feature concatenation to transformer-based cross-modal attention, within a unified framework that leverages pretrained modality-specific encoders, enabling a principled evaluation of how fusion depth affects behavior prediction performance.
Existing studies on wearable-based behavior recognition are often constrained by lab-based validation, limited participant diversity, and a focus on retrospective detection rather than prediction of behaviors \cite{francese_supporting_2022,shamhan_advancements_2025}.
Meanwhile, recent advances in foundation models trained on large-scale wearable datasets—such as the Pretrained Actigraphy Transformer (PAT) and cross-modal encoders integrating accelerometry and photoplethysmography (PPG)—have achieved state-of-the-art results in domains including sleep staging, mood prediction, and activity classification \cite{ruan_ai_2025,yuan_self-supervised_2024,abbaspourazad_wearable_2025}.
Prior multimodal wearable studies in autism have primarily focused on retrospective classification and detection tasks conducted in controlled or laboratory settings, with limited validation in real-world environments \cite{francese_supporting_2022,welch_use_2022,shamhan_advancements_2025}.
Additionally, many systems emphasize post hoc behavior recognition rather than early prediction for proactive intervention \cite{rad_motion_2025,zwilling2022prediction}.
To address these limitations, our study developed a behavior prediction model for multimodal signals—accelerometry, electrodermal activity, and skin temperature—collected in real-world special education classrooms.
By enabling early prediction of challenging behaviors, our approach moves beyond passive monitoring toward scalable, proactive intervention systems tailored to the behavioral complexity of profound autism.

\section{Method and Material}

\subsection{Data Collection and Behavior Annotation}\label{sec:data_collect}

The study was conducted at The Center for Discovery (TCFD), providing residential programs for individuals with multiple disabilities and autism.
Nine male residents, aged between 10 and 21 years, participated in the study (see Table~\ref{tab:participants}).
Institutional Review Board (IRB) approval was obtained from TCFD, and informed consent was secured from the parents or guardians of each participant.
Behavioral monitoring and annotation were performed by trained behavior analysts and research assistants, who defined and identified specific behaviors of interest for each participant.
Video annotation was conducted using Noldus The Observer XT 15, which labeled each instance of a target behavior.

During class activities, students wore Q-Sensor bands (Affectiva Inc., Cambridge, MA), which collect accelerometer (30 Hz), electrodermal activity (EDA; 30 Hz), and skin temperature (Tsk; 30 Hz) time-series data.
Although accelerometer signal morphology can differ substantially between wrist and ankle due to varying movement dynamics, EDA signals, being driven by eccrine sweat gland activity, remain interpretable at both on-body locations, though signal amplitude may differ \cite{hernandez-capistran_commercial_2024}.
Importantly, EDA was used as the primary inclusion criterion for data analysis.
A 5-second analysis window was included only if the EDA signal demonstrated valid skin contact and physiological variability.
Sessions, defined as daily recordings, were excluded if the EDA signal exhibited common artifacts such as flat-line responses (no phasic or tonic change) or abrupt signal dropout.
The rationale for using EDA validity as a gating criterion is two folds: (1) prior work has identified EDA as a robust correlate of arousal and agitation in individuals with autism \cite{zhu_stress_2023,welch_use_2022}, and (2) signal integrity in EDA is more sensitive to proper sensor placement and skin contact compared to accelerometry, making it a practical marker for session-level data quality screening.

\begin{figure}[t]
    \centering
    \begin{subfigure}[b]{0.40\linewidth}
        \includegraphics[width=\linewidth]{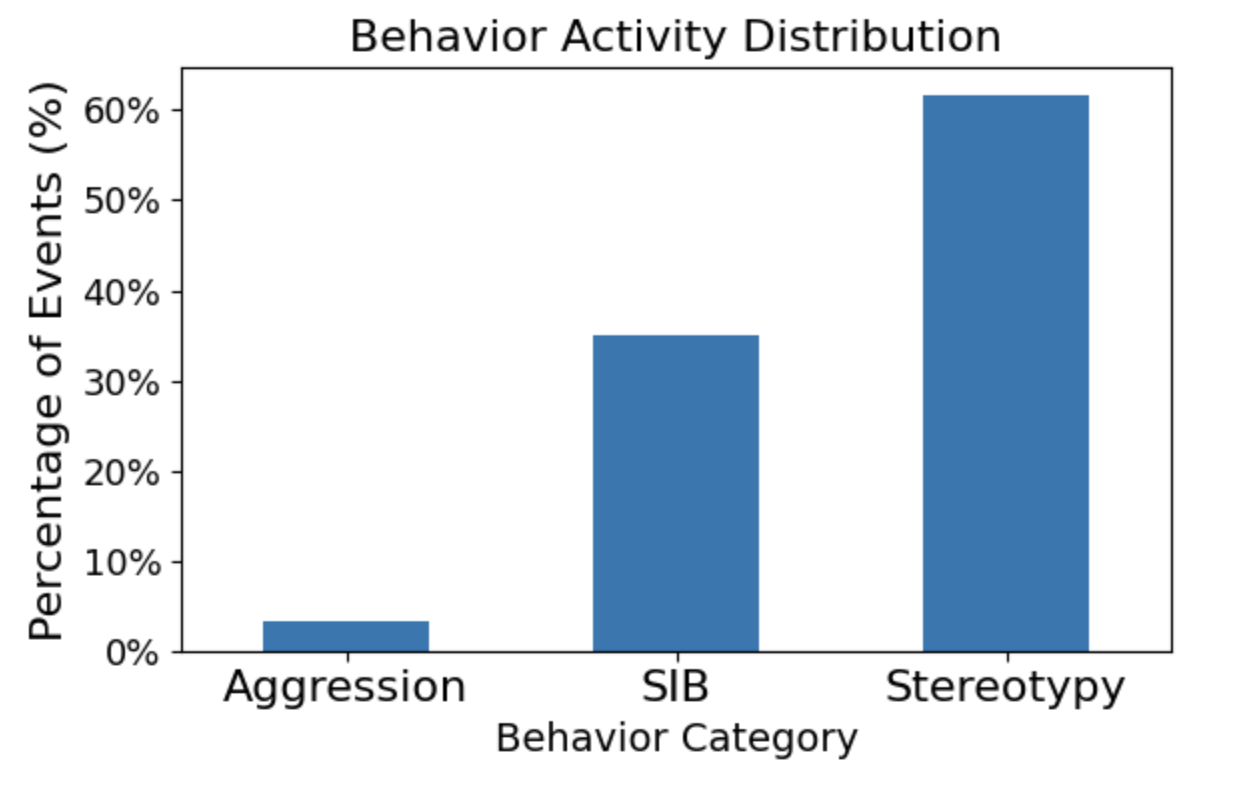}
        \caption{Total percentage events of target behaviors.}
        \label{fig:behavior_duration}
    \end{subfigure}
    \hfill
    \begin{subfigure}[b]{0.58\linewidth}
        \includegraphics[width=\linewidth]{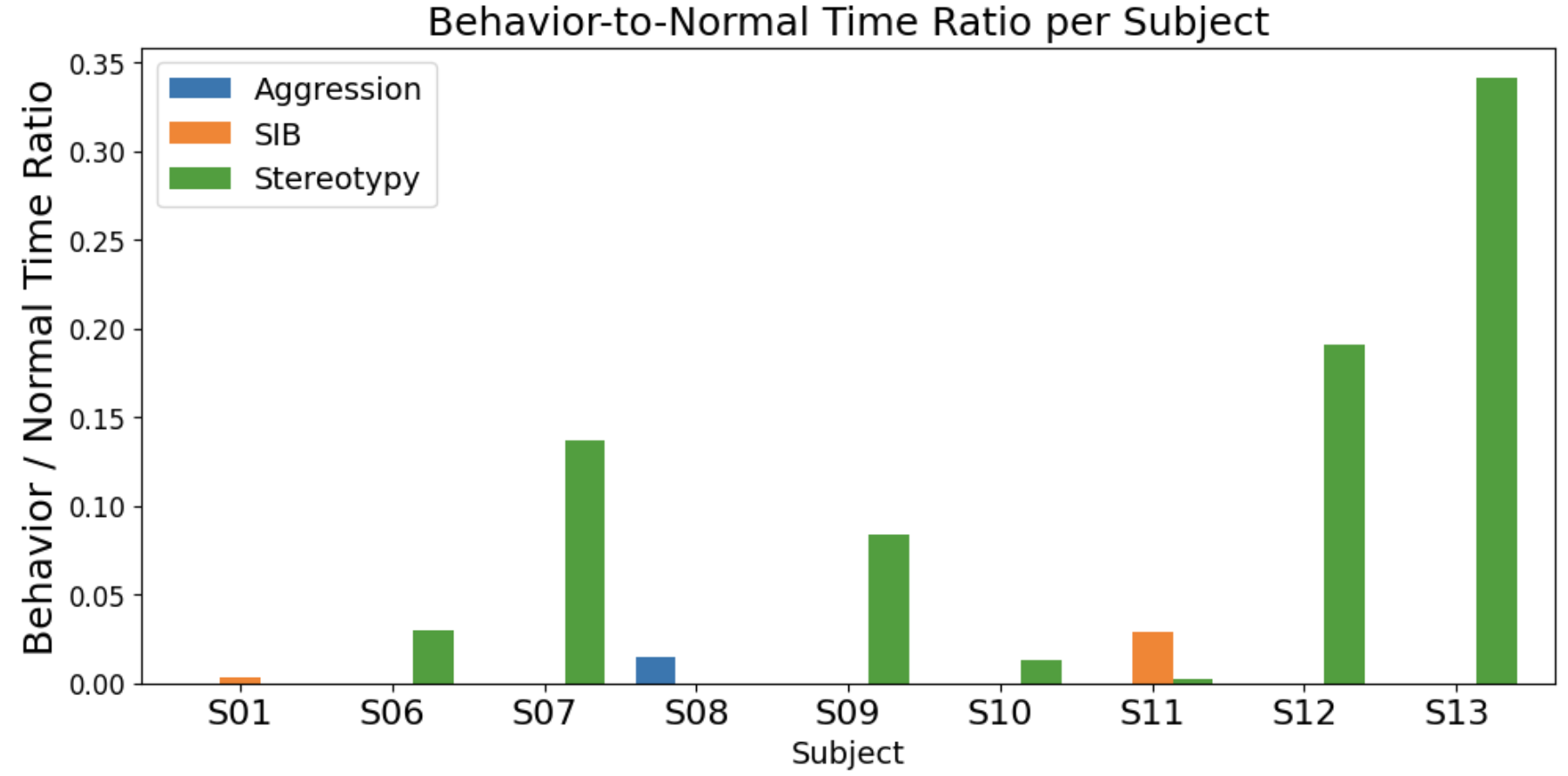}
        \caption{Total behavior duration ratio to no behavior per subject (in minutes).}
        \label{fig:subject_duration}
    \end{subfigure}
    \caption{Distribution of target behaviors in the labeled dataset. (a) shows the behavior distribution. (b) shows ratio of the total ASD behavior duration per subject to normal behavior.}
    \label{fig:combined_behavior}
\end{figure}

Each student's wearable sensor time series was labeled with challenging episodes, including motor stereotypies, aggression, and self-injurious behavior (SIB).
The total duration of labeled wearable time-series data was 110.7 hours.
Table~\ref{tab:participants} summarizes subject demographics, number of recording sessions, recorded modalities, sensor location, and total hours of labeled data per subject.
All recordings included synchronized accelerometry, electrodermal activity, and skin temperature signals.
\autoref{fig:behavior_duration} shows the distribution of behavior activity categories: stereotypy was the most frequently observed class (over 1,200 events), followed by SIB (approximately 700 events), and aggression (100 events), which was comparatively rare.
This class imbalance reflects behavioral expression patterns common in profound autism and motivates the use of imbalance-aware learning strategies.
\autoref{fig:subject_duration} presents the subject-wise duration of labeled behavioral episodes, illustrating inter-individual variability in both frequency and type of challenging behavior.
Some participants demonstrated prolonged stereotypy episodes, while others exhibited only brief instances of aggression or SIB.

The study focused on eight individually operationalized challenging behaviors encompassing a broad range of actions: (1) motor stereotypies, (2) aggression, (3) self-injurious behavior, (4) self-injurious jump, (5) hand bite, (6) dropping, (7) jumping, and (8) disruptive behavior.
To improve model robustness and reduce label sparsity, these behaviors were grouped into three high-level behavior bins based on behavioral similarity and clinical relevance: (1) Aggression, (2) Self-Injurious Behavior, and (3) Stereotypy.
This binning was performed in two stages.
First, raw annotation labels were mapped to standardized secondary bins, for instance, ``self-injurious jump'' and ``hand bite'' were grouped under ``self-injurious behaviors,'' while ``dropping'' and ``jumping'' were combined under ``Stereotypy behavior.'' 
In the final step, higher-level bins were defined: all aggression-related behaviors were grouped under ``Aggression,'' all self-injurious behaviors under ``SIB,'' and repetitive motor behaviors under ``Stereotypy.''
This hierarchical binning strategy enabled effective multi-class modeling while preserving the clinical distinctiveness of challenging behavior types.

\begin{table}[t]
\centering
\caption{Dataset used in this study.}
\label{tab:participants}
\begin{tabular}{lccccc}
\hline
\textbf{Subject ID} &
\textbf{\begin{tabular}[c]{@{}c@{}}Age range\\ (years)\end{tabular}} &
\textbf{\# of sessions} &
\textbf{Sensor location} &
\textbf{\begin{tabular}[c]{@{}c@{}}Total duration of\\ sessions (h)\end{tabular}} \\ \hline
S01 & 17--20 & 23 & Right ankle & 21.4 \\
S06 & 14--16 & 6  & Left ankle  & 5.1 \\
S07 & 10--13 & 23 & Left wrist  & 14.6 \\
S08 & 16--17 & 12 & Left ankle  & 12.8 \\
S09 & 12--14 & 24 & Left ankle  & 20.3 \\
S10 & 14--15 & 8  & Left ankle  & 8.1 \\
S11 & 14--15 & 22 & Left ankle  & 12.8 \\
S12 & 20--21 & 12 & Right ankle & 10.8 \\
S13 & 19--20 & 6  & Left ankle  & 4.8 \\ \hline
\end{tabular}
\end{table}

\subsection{Multimodal Sensor Time Series Analysis for Challenging Behavior Detection and Prediction}\label{sec:approach}

\subsubsection{Overall Analysis Pipeline}

The overall pipeline integrates three physiological modalities—acceleration, electrodermal activity, and skin temperature—to detect or predict challenging behavior episodes in students with profound autism.
Our approach follows the general structure of the human activity recognition pipeline \cite{bulling_har}, consisting of temporal segmentation, feature extraction, multimodal fusion, and classification.
Each raw modality-specific time series is first segmented using a sliding window approach, enabling frame- or window-wise modeling of temporal dynamics.
Next, modality-specific encoders are used to extract spatiotemporal representations tailored to the signal characteristics of each input stream.
These representations are then aligned and aggregated through a multimodal fusion mechanism that supports both intra-modality and inter-modality temporal dependencies.
In our implementation, we explore multiple fusion strategies, including hierarchical and attention-based methods, to integrate complementary information across modalities.
The resulting fused embeddings are used as input to a prediction head that performs binary or multiclass classification, depending on the task.
This design allows for flexible experimentation with unimodal baselines, fusion of features from unimodal baselines, and feature-level fusion approaches, enabling a comprehensive evaluation of multimodal wearable sensing strategies for behavior detection and prediction in real-world special education settings.

\begin{figure}[t]
\centering
\subcaptionbox{Modality-specific}{%
    \includegraphics[width=0.8\textwidth]{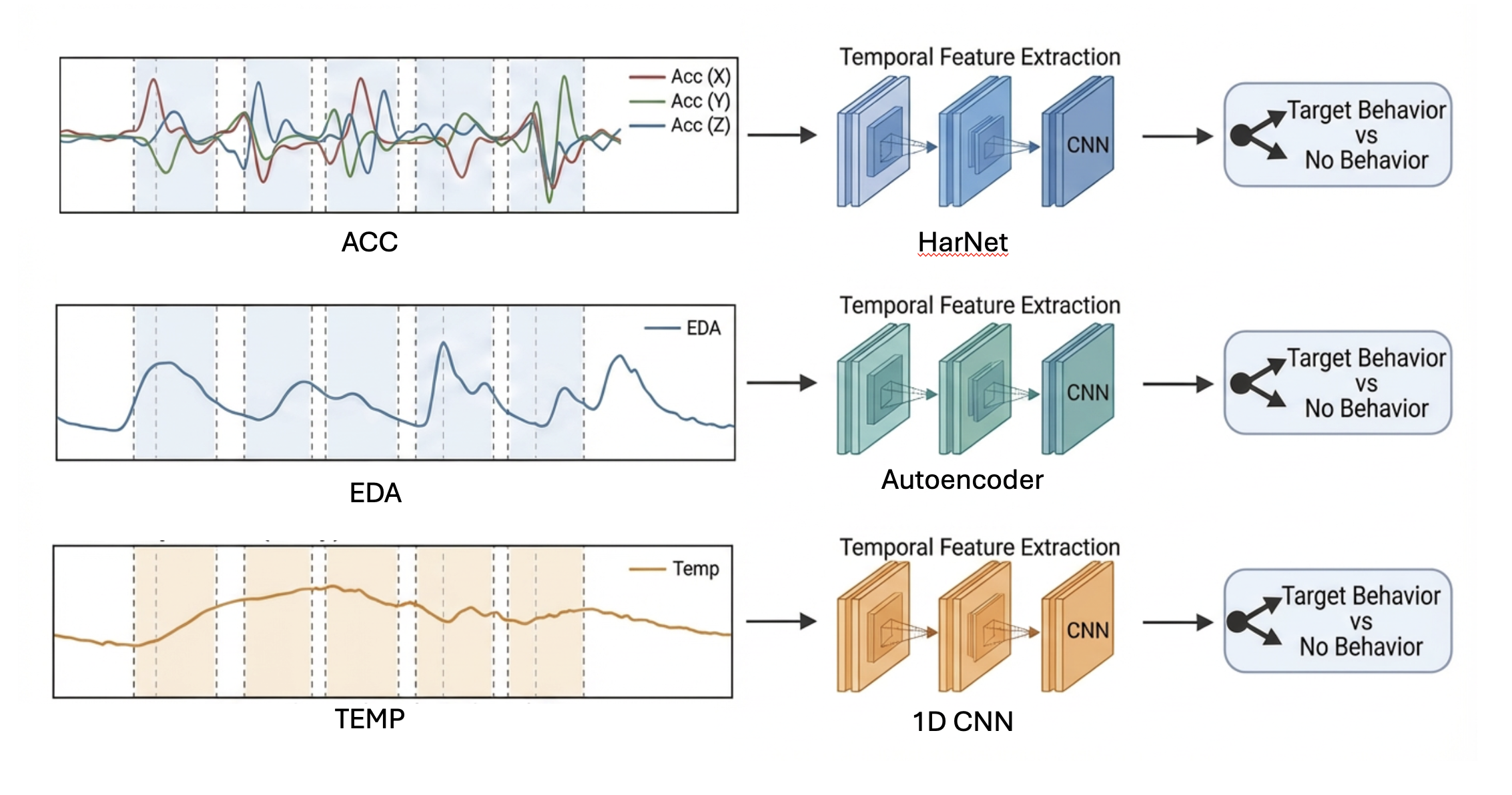}}
    \\
\subcaptionbox{Naive feature concatenation}{%
    \includegraphics[width=0.8\textwidth]{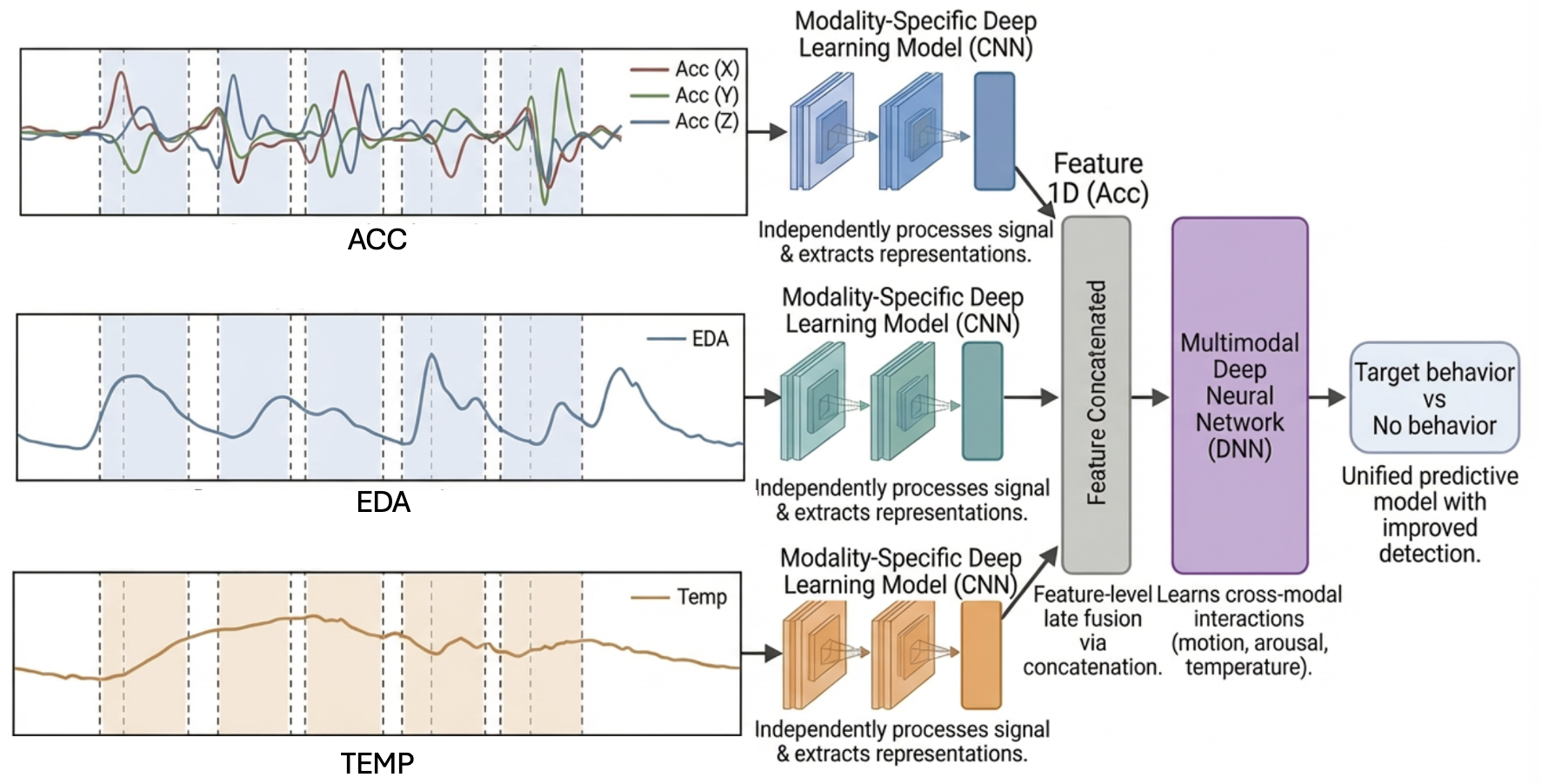}}
\caption{Model architecture for (a) modality-specific analysis and (b) naive multimodal fusion through temporal pooling, feature concatenation, and multi-layer perceptron.
}
\label{fig:arch_s}
\end{figure}

\begin{figure}[t]
\centering
\subcaptionbox{Temporal self-attention}{%
    \includegraphics[width=0.8\textwidth]{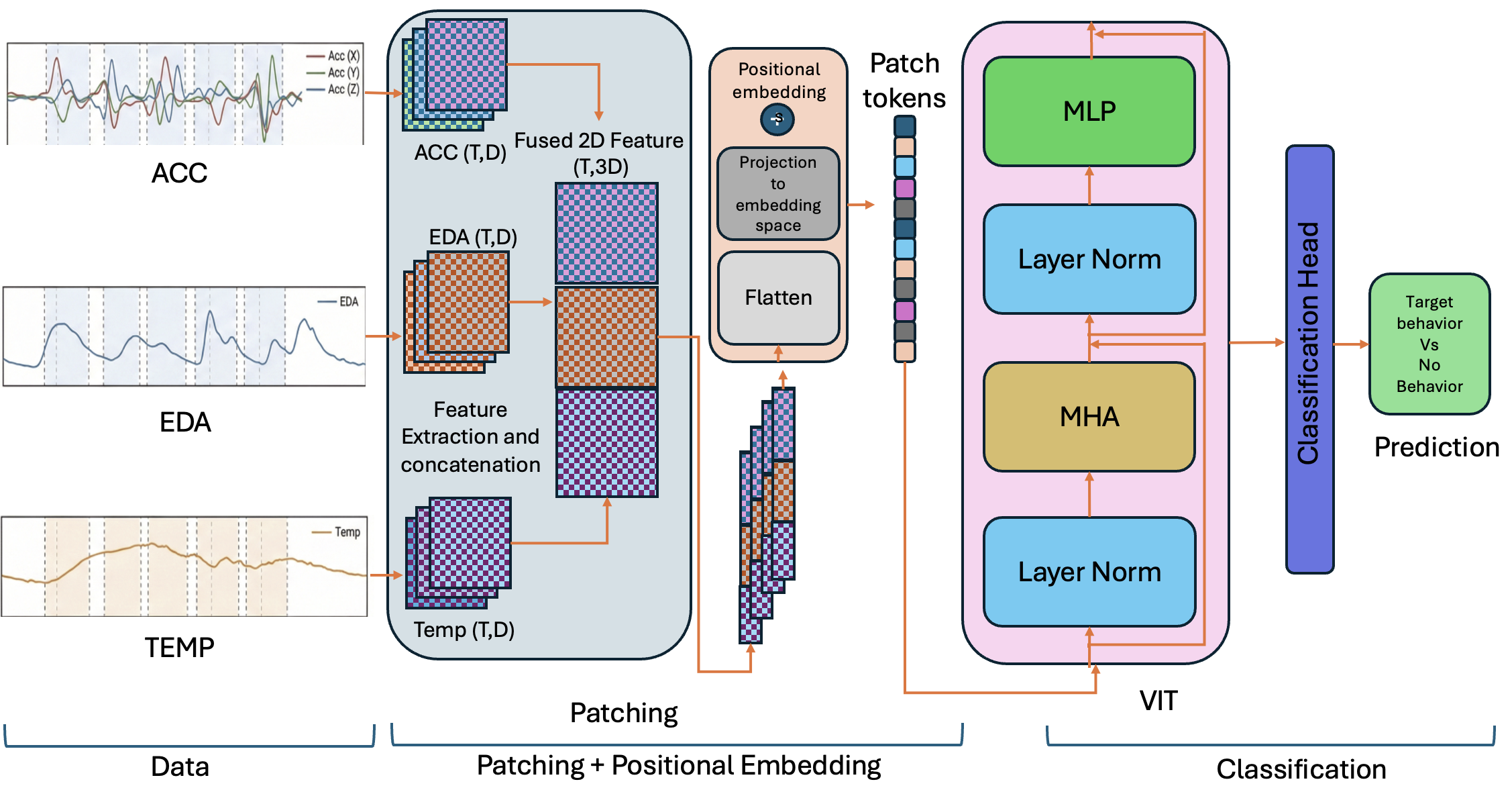}}
    \\
\subcaptionbox{Cross-time cross-modal attention}{%
    \includegraphics[width=0.8\textwidth]{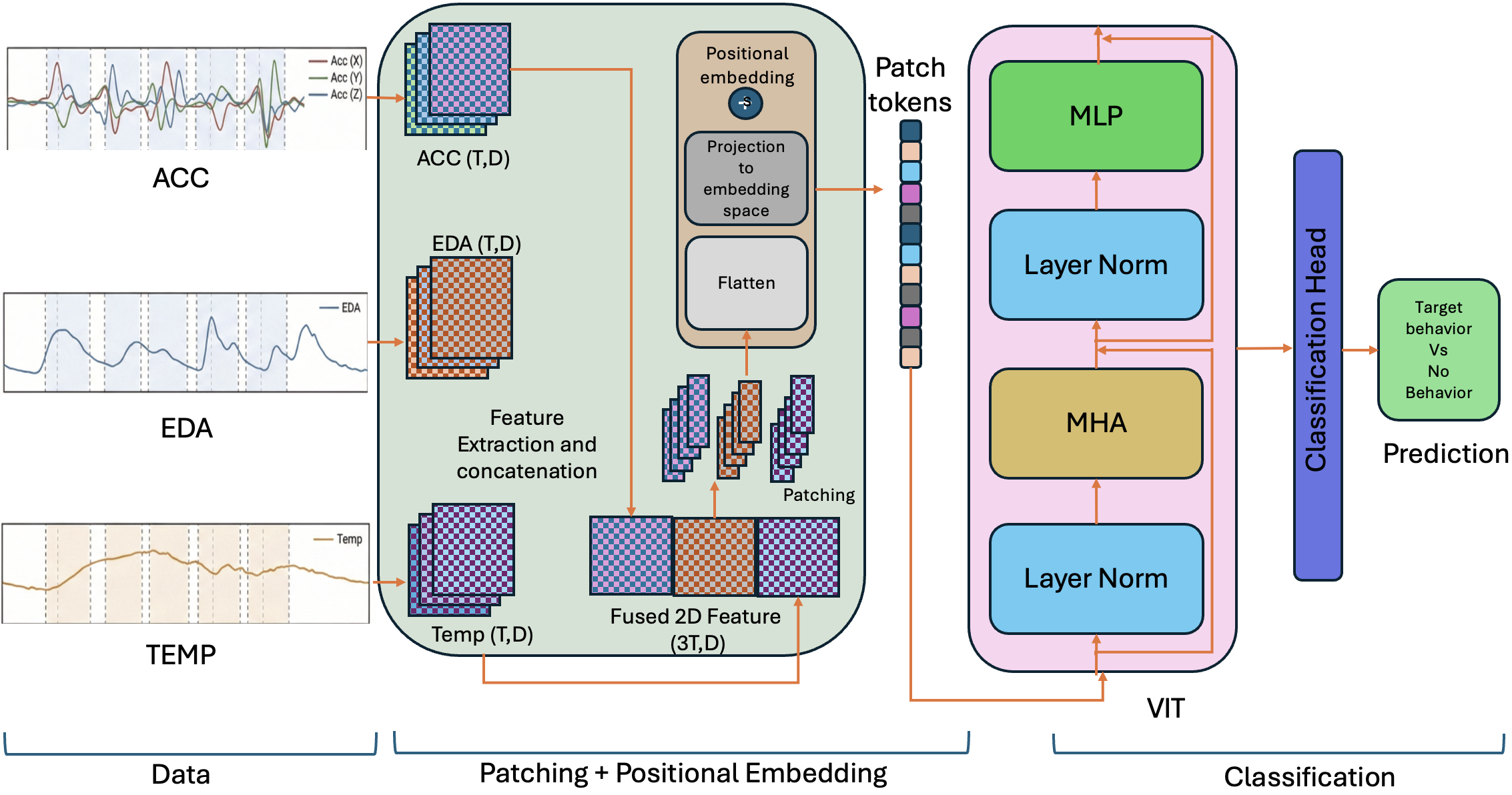}}
\caption{Model architecture for (a) temporal self-attention of concatenated multimodal features aligned in time using vision Transformer (ViT) and
(b) spatiotemporal multimodal cross-attention using ViT.}
\label{fig:arch_vit}
\end{figure}

\subsubsection{Data Preprocessing}

The raw Q-Sensor recordings provide time-synchronized multimodal time series consisting of tri-axial acceleration, electrodermal activity, and skin temperature.
To prepare these signals for modeling, modality-specific preprocessing was applied to enhance signal quality, standardize scales, and align behavioral labels with sensor data.
First, the acceleration and skin temperature channels were scaled to the range $[0,1]$ using min–max normalization to account for differences in measurement units and magnitude.
EDA signals were decomposed into tonic and phasic components using the \texttt{NeuroKit2} library \cite{Makowski2021neurokit}, which implements convex optimization methods to separate slow-varying baseline activity from rapid, event-related fluctuations.
For downstream analysis, we retained only the tonic component, as it represents overall sympathetic arousal and reduces sensitivity to transient noise that can arise from abrupt motion or electrode artifacts \cite{boucsein_electrodermal_2012}.
Time-synchronized multimodal time series data were divided into temporal segments using a sliding window of 5 seconds, as this provides sufficient temporal context to capture characteristic movement and physiological patterns associated with challenging behaviors such as aggression, motor stereotypy, and self-injury \cite{anumula_understanding_physiological_2025}.
A sliding window strategy with a stride of one second was employed to increase the number of positive training samples and ensure adequate temporal coverage of challenging behavior episodes.
This approach partially mitigates class imbalance by enabling multiple overlapping windows to capture different phases of the same behavioral event, including onset, peak, and offset.
While sliding-window segmentation is common in time-series classification \cite{anumula_understanding_physiological_2025}, we selected a fixed stride to balance class representation and data efficiency without introducing additional complexity such as adaptive windowing schemes \cite{adapslide}.
Each window inherits a challenging behavior label if it overlaps with any manually annotated episode of motor stereotypy, aggression, or self-injurious behavior from the video ground truth (see Sec.~\ref{sec:data_collect}).
This procedure ensured that even brief or transitional behavioral events were captured during segmentation, resulting in a fully synchronized and behavior-aligned multimodal dataset.

\subsubsection{Modality-specific feature representations}\label{sec:multimodal_fusion}

We explored various approaches to extract behavioral features from each sensor time series by adapting state-of-the-art models, as shown in \autoref{fig:arch_s}a.

\paragraph{Accelerometer}

We evaluated three classes of deep learning models for extracting and modeling accelerometer features: HarNet5 \cite{yuan_self-supervised_2024}, DeepConvLSTM \cite{deepconv}, and a Transformer-based classifier \cite{ruan_ai_2025}.
HarNet5 is a ResNet-style 1D convolutional architecture pretrained on the large-scale Capture-24 dataset \cite{chan_capture24_2024}, comprising over 3,000 hours of wearable data via self-supervised representation learning \cite{yuan_self-supervised_2024,abbaspourazad_wearable_2025,ruan_ai_2025}.
In our pipeline, we fine-tuned the pretrained HarNet5 backbone.
Specifically, pretrained weights were loaded for all convolutional layers, and the final classification head was replaced with a task-specific layer matching our binary or multiclass outputs.
All layers were jointly updated during supervised training on the labeled ASD accelerometer segments.
This strategy leverages general movement and activity representations from the foundation model while adapting them to challenging behavior prediction.
For DeepConvLSTM \cite{deepconv}, we adapted a hybrid architecture that combines convolutional and recurrent modeling, which has been widely adopted in human activity recognition literature.
The model consists of five stacked 1D convolutional layers for local temporal feature extraction, followed by two long short-term memory (LSTM) layers to capture longer-range temporal dependencies in the accelerometer sequence.
To ensure a fair comparison with pretrained foundation models, DeepConvLSTM was first pretrained on the large-scale Capture-24 dataset using supervised activity labels and subsequently fine-tuned on our ASD behavior dataset using the same input preprocessing and windowing strategies.
During fine-tuning, all layers were updated end-to-end, and the final classification layer was adapted to the target challenging behavior labels.
This staged training strategy enables the model to benefit from generalizable movement representations learned from population-scale data while adapting to the unique behavioral patterns observed in individuals with profound autism.
The Transformer-based classifier was designed to capture long-range temporal dependencies in accelerometer sequences.
We adopted the Actigraphy Transformer architecture proposed by Ruan \textit{et al.}~\cite{ruan_ai_2025}, which was pretrained using self-supervised learning on the Capture-24 dataset.
The input 5-second accelerometer windows were divided into temporal patches of 15 samples and projected into embedding space, followed by the addition of learnable positional encodings.
The embeddings were passed through a stack of transformer encoder layers, each containing multi-head self-attention and feedforward sublayers with residual connections and layer normalization.
For this study, we fine-tuned the pretrained encoder on our ASD dataset by replacing the final classification head with a task-specific dense layer suited for challenging behavior classification.

\paragraph{Electrodermal Activity}

To model electrodermal activity, we used a convolutional autoencoder-based encoder approach \cite{yu_semi-supervised_2023} to learn temporal representations from tonic EDA signals.
In parallel, a handcrafted feature-based classifier \cite{kliangkhlao_electrodermal_2024} was used as a baseline to evaluate the discriminative capacity of EDA in detecting challenging behavioral episodes.
While the autoencoder served as the main modeling strategy, the feature-based approach allowed us to test the standalone utility of EDA for behavior classification.
The convolutional autoencoder was adapted from \textit{Yu et al.}~\cite{yu_semi-supervised_2023} and pretrained on the WESAD dataset \cite{schmidt_wesad_2018}, a widely used multimodal stress dataset.
The encoder consisted of five 1D convolutional layers with channel dimensions [4, 8, 16, 32, 64], each followed by batch normalization, ReLU activation, and max-pooling layers with stride 2.
The decoder mirrored the encoder using transposed convolutions to reconstruct the original signal.
The final encoder output was averaged using global average pooling to obtain a compact latent feature representation.
We used tonic EDA as input and fine-tuned the encoder weights on our ASD dataset for supervised classification.
For the shallow machine learning baseline, we implemented a handcrafted feature model inspired by previous work \cite{zhu_feasibility_2022,kliangkhlao_electrodermal_2024}.
We extracted tonic EDA descriptors over fixed 5-second windows, including (1) number of peaks obtained via prominence thresholding, (2) maximum amplitude, (3) mean conductance, and (4) standard deviation.
These features were z-normalized and input to a $k$-nearest neighbors (KNN) classifier.
This approach was selected due to its simplicity, interpretability, and effectiveness for small datasets, as supported by prior studies in physiological signal classification \cite{zhu_feasibility_2022}.

\paragraph{Skin Temperature}

We explored two different modeling strategies for skin temperature analysis: handcrafted feature extraction and a convolutional neural network (CNN) model.
The handcrafted feature with shallow model approach followed prior works \cite{le_tran_thuan_machine_2024,logacjov_machine_2024} to capture coarse thermal trends related to stress or emotional states in wearable sensing.
We computed descriptive statistics over 5-second windows, including mean temperature, standard deviation, and slope-based metrics.
These features were input to a $k$-nearest neighbors classifier.
This baseline provides interpretable features with minimal computational cost.
To capture richer temporal dynamics, we implemented a lightweight 1D CNN.
The model used three 1D convolutional layers with kernel sizes 7, 5, and 3, ReLU activations, and channel depth increasing from 16 to 64.
The model was trained on a later released temperature time-series data from Capture-24 \cite{chanchan2023capture24} using the Adam optimizer with a learning rate of $1\times10^{-4}$  and a batch size of 16, then fine-tuned to our dataset.

\subsubsection{Multimodal Fusion}

To integrate information across heterogeneous modalities, we designed a mid-fusion multimodal framework combining the latent representations of acceleration, electrodermal activity, and skin temperature.
As shown in \autoref{fig:arch_s}b and \autoref{fig:arch_vit}a,b, we explored three fusion strategies: (i) naive feature concatenation of temporally pooled multimodal features, (ii) temporal self-attention over time-aligned multimodal embeddings, and (iii) cross-time cross-modality attention over multimodal time-series embeddings.
In the naive fusion approach (\autoref{fig:arch_s}b), we first applied temporal pooling (e.g., mean or max) to each modality’s feature sequence, resulting in fixed-size vectors (acceleration $d_{\text{ACC}}=128$, EDA $d_{\text{EDA}}=64$, skin temperature $d_{\text{Tsk}}=64$).
These pooled representations were concatenated and passed through a multi-layer perceptron (MLP) with three hidden layers of 256 nodes and ReLU activations.
This baseline enables fast and lightweight multimodal integration but does not explicitly model temporal or cross-modal interactions.

We also explored transformer-based mid-fusion architectures to better capture both multimodal and temporal dependencies. Specifically, we evaluated two variants that differ in how they attend and integrate signals across modalities and time. 
The temporal self-attention model learns the cross-time attention on time-aligned multimodal timeseries, assuming that each modality has similar response rates (\autoref{fig:arch_vit}a).
$T$ timesteps of features from modality-specific encoders (accelerometer $(T,d_{\text{ACC}})$, EDA $(T, d_{\text{EDA}})$, and skin temperature $(T, d_{\text{Tsk}})$ are then stacked along the feature axis (Depth-wise concatenation) to form a single sequence of shape \( (T, D) \), where $D = d_{\text{ACC}} + d_{\text{EDA}} + d_{\text{Tsk}}$. 
This sequence is patchified into non-overlapping patches of 10 time steps \( (10, D) \) and passed into a Vision Transformer (ViT), enabling it to learn temporal patterns in time-aligned multimodal time-series. 
On the other hand, the cross-temporal cross-modal attention model learns the cross-modality temporal attention and cross-modal attentions simultaneously (\autoref{fig:arch_vit}b).
We first upscale EDA ($d_{\text{EDA}}=64$), Tsk ($d_{\text{Tsk}}=64$) to 128 through a linear projection layer, then concatenate three modality features with $(T, d=128)$ along the time axis, $(3T, d)$.
Patches are then extracted along the temporal axis (e.g., 10 time steps per patch, $(10,d)$) and processed by the ViT. This configuration encourages the model to focus on cross-modal interactions as well as on different time steps. 
Together, these two designs allow us to compare the effects of different fusion strategies: one emphasizing temporal dependencies only, and the other emphasizing cross-time and cross-modality at the same time.

This mid-fusion modeling strategy is motivated by known physiological dynamics of wearable biosignals.
EDA reflects sympathetic nervous system arousal via sweat gland activity, and canonical descriptions note that EDA responses, particularly phasic skin conductance responses, are not instantaneous but exhibit measurable latency relative to external stimuli, often on the order of one to several seconds \cite{boucsein_electrodermal_2012,setz_discriminating_2010}.
Such delays occur because eccrine sweat gland activation and subsequent changes in skin conductance require time to develop following a stressor or cognitive event, in contrast to more immediate accelerometer signals linked directly to motion.
Skin temperature responds even more slowly, reflecting peripheral vasoconstriction or vasodilation driven by autonomic regulation rather than rapid sympathetic spikes, and thermal changes typically lag both motion and EDA due to slower vascular and heat transfer processes \cite{shin2025skintemp}.
Together, these characteristics imply that physiological cues (EDA and temperature) may be temporally misaligned with movement patterns, motivating transformer models that can attend across time and modality to capture delayed and distributed multimodal signatures for behavior detection.
For both ViT-based approaches, the class token (\textit{cls}) is used for behavior detection and prediction.

\subsection{Experimental Settings and Evaluation Protocol} \label{sec:exp}

We assumed that future behavior prediction, where the model must capture precursors of motion and physiological cues, is more challenging than concurrent detection.
We also considered four-way behavior classification (Aggression, Self-Injurious Behavior (SIB), Stereotypy, and None) more challenging than binary classification of challenging versus non-challenging behavior, as multiclass recognition requires modeling fine-grained behavioral distinctions. 
Because some subjects exhibited only a single behavior type, subject-wise five-fold cross-validation was infeasible for four-class classification; therefore, we used four-fold cross-validation for that setting.
We opted for subject-wise five-fold cross-validation over leave-one-subject-out (LOSO) evaluation to ensure that each test fold contained multiple subjects, enabling assessment of model performance across diverse behavioral profiles within each evaluation round.
With only nine participants exhibiting heterogeneous behavior types and frequencies, LOSO would yield single-subject test sets whose metrics reflect individual behavioral characteristics rather than generalizable model performance, and would preclude meaningful within-fold variance estimation.
As a pathway toward behavior prediction, we adopted a step-by-step approach progressing from simpler tasks (e.g., unimodal and binary detection) to more complex tasks (e.g., multimodal and multiclass prediction), allowing outcomes at each stage to inform optimization of multimodal time-series analysis.
We first evaluated unimodal and multimodal fusion pipelines for binary detection of challenging behavior across the design options described in Sec.~\ref{sec:approach}.
The selected pipeline was then adapted to predict challenging episodes up to 30 minutes ahead and to perform multiclass detection and prediction.
We further compared two multiclass prediction configurations: a four-class formulation (None, Aggression, SIB, and Stereotypy) and a coarser three-class grouping that combined Aggression and SIB into a single high-risk class (high-risk, low-risk stereotypy, and no behavior) to assess performance across risk levels.
To validate generalizability, all experiments except the four-class experiment were conducted using five runs of subject-wise five-fold nested cross-validation, with 60\%, 20\%, and 20\% of subjects allocated to training, validation, and test sets, respectively.
We reported F1 score, AUC-ROC, precision, and recall with 95\% confidence intervals.
All deep learning experiments were implemented in \texttt{PyTorch} and trained for 200 epochs using the Adam optimizer with a base learning rate of $1\times10^{-4}$ and weight decay of $1\times10^{-5}$.
Learning rates were layer-specific, with smaller rates ($1\times10^{-5}$) for pretrained backbones (HarNet5 , CNN 1D and the EDA autoencoder).
The batch size was set to 64.
Given severe class imbalance, where non-behavior windows outnumber behavior windows, we applied batch-level label balancing during training.
normal behavior windows (labeled 0 or negative sample) were randomly subsampled within each batch to maintain a fixed label ratio of 1.5:1 (label~0:label~1), increasing exposure to rare events with ASD behaviors (labeled 1 or positive sample) while preserving contextual diversity.
Balancing was applied only to the training set, while validation and test sets retained natural class distributions for unbiased evaluation.
Hyperparameters, including learning rate, segment length (1–7 seconds), and label ratio, were selected via manual grid search based on cross-validation stability and held fixed across experiments for reproducibility.

To evaluate the feasibility of predicting challenging behaviors at varying lead times, we conducted a temporal horizon analysis by training models to predict future behavior events at intervals ranging from 30 seconds to 1800 seconds (30 minutes).
This was implemented by offsetting behavior labels forward in time by the target prediction window, allowing the model to learn patterns that precede the onset of challenging behaviors.
To interpret the relative contribution of each modality to behavior prediction, we applied MM SHAP (SHapley Additive exPlanations) analysis \cite{Parcalabescu_2023} to our trained multimodal models for a 10-minute prediction task, which represented the maximum horizon at which prediction performance remained comparable to detection performance.
We aggregated feature attributions at the modality level to quantify the contribution of acceleration, electrodermal activity, and skin temperature.
To interpret which temporal windows of the input time series most influenced the model’s predictions, we applied Grad-CAM (Gradient-weighted Class Activation Mapping) \cite{Selvaraju_2019} to the final convolutional layer of the multimodal model for the 10-minute prediction task.
As Grad-CAM produces instance-level activation maps, a randomly selected high-confidence positive-class sample was used for visualization to avoid selection bias while ensuring the depicted prediction corresponded to a clinically relevant behavioral episode.


\section{Results}

We evaluate our multimodal framework across all subjects using multiple runs. These include detection and prediction results of challenging behaviors.
Table~\ref{tab:model-config} summarizes the hyperparameters of models selected in our experiment.

\subsection{Detecting Challenging Behaviors in Profound Autism}

\begin{table*}[t]
\centering
\caption{Final model architecture and hyperparameter configuration used for all reported experiments.}
\label{tab:model-config}
\resizebox{\textwidth}{!}{%
\begin{tabular}{|l|l|}
\hline
\textbf{Component} & \textbf{Configuration} \\
\hline
\textbf{Accelerometer (ACC)} & HarNet5~\cite{yuan_self-supervised_2024}, pretrained transformer classifier~\cite{ruan_ai_2025}, DeepConvLSTM~\cite{deepconv} fine-tuned \\
\textbf{EDA Encoder} & CNN autoencoder~\cite{yu_semi-supervised_2023}, pretrained (WESAD), encoder fine-tuned \\
\textbf{Temperature Encoder} & 1D CNN: 3 conv layers (kernel sizes 7, 5, 3), 64 channels, ReLU, adaptive avg pooling, proj to 128-dim \\
\hline
\textbf{ Naive Fusion (MLP)} & Temporal pooling + feature concatenation + 3-layer MLP (256 units, ReLU) \\
\textbf{ Temporal Self-Attention (ViT)} & Input: $(3{\times}128,150)$, Patch: $(3{\times}128,10)$, Dim=128, Depth=2, Heads=4, MLP Dim=256 \\
\textbf{ Spatiotemporal Cross-modal Attention (ViT)} & Input: $(128,154)$, Patch: $(128,11)$, Dim=128, Depth=2, Heads=4, MLP Dim=256 \\
\hline
\textbf{Optimizer} & Adam \\
\textbf{Batch Size} & 64 \\
\textbf{Epochs} & 200 \\
\textbf{Learning Rates} & HarNet5: $1\times10^{-5}$, EDA encoder: $1\times10^{-5}$, Temp/Classifier: $1\times10^{-4}$ \\
\textbf{Weight Decay} & $1\times10^{-5}$ \\
\hline
\textbf{Segment Length} & 150 samples (5 seconds) \\
\textbf{Label Balancing Ratio} & 1.5 (undersampling majority class during training) \\
\textbf{Normalization} & Per-channel min-max normalization (computed from train fold only) \\
\hline
\end{tabular}%
}
\end{table*}

\subsubsection{Per-modality behavior detection model}

Table~\ref{tab:per_modality_metrics} summarizes the performance of different models trained on individual modalities. For accelerometer-based model, HarNet model achieved the highest AUC-ROC of 77.8\% when compared to other models.
For EDA and temperature, deep learning-based models showed 62.0\% and 62.4\% AUC-ROC, outperforming the handcrafted features, consistent with findings from a previous study \cite{zhu_feasibility_2022}.
Among the three modalities, the accelerometer-based model produced the strongest discriminative ranking (highest AUC-ROC).
The best deep learning models for each modality were used for multimodal fusion analysis.

\begin{table}[t]
  \caption{Challenging behavior detection performance of per-modality models. Best and second-best scores per metric are \textbf{bolded} and \underline{underlined}, respectively.}
  \label{tab:per_modality_metrics}
  \centering
  \begin{tabular}{llcccc}
    \toprule
    \textbf{Modality} & \textbf{Model} & \textbf{AUC-ROC} & \textbf{Precision} & \textbf{Recall} & \textbf{F1 Macro} \\
    \midrule
    \multirow{3}{*}{Acceleration}
    & \textbf{HARNET}~\cite{yuan_self-supervised_2024} & \textbf{0.778} $\pm$ 0.089 & {0.276} $\pm$ 0.134 & 0.595 $\pm$ 0.144 & \underline{0.608} $\pm$ 0.084 \\
    & DeepConvLSTM~\cite{deepconv} & \underline{0.656} $\pm$ 0.047 & 0.170 $\pm$ 0.047 & 0.494 $\pm$ 0.106 & 0.517 $\pm$ 0.037 \\
    & Transformer~\cite{ruan_ai_2025} & 0.654 $\pm$ 0.063 & 0.140 $\pm$ 0.042 & {0.729} $\pm$ 0.136 & 0.415 $\pm$ 0.041 \\
    \midrule
    \multirow{2}{*}{EDA}
    & Handcrafted~\cite{zhu_feasibility_2022} & 0.575 $\pm$ 0.026 & 0.149 $\pm$ 0.047 & 0.039 $\pm$ 0.005 & 0.494 $\pm$ 0.012 \\
    & \textbf{Conv. autoencoder}~\cite{yu_semi-supervised_2023} & 0.620 $\pm$ 0.049 & \underline{0.586} $\pm$ 0.025 & \textbf{0.861} $\pm$ 0.095 & 0.604 $\pm$ 0.035 \\
    \midrule
    \multirow{2}{*}{Skin Temp.}
    & Handcrafted~\cite{le_tran_thuan_machine_2024, logacjov_machine_2024} & 0.542 $\pm$ 0.022 & 0.140 $\pm$ 0.054 & 0.074 $\pm$ 0.029 & 0.502 $\pm$ 0.019 \\
    & \textbf{CNN-based model} & 0.624 $\pm$ 0.049 & \textbf{0.588} $\pm$ 0.036 & \underline{0.854} $\pm$ 0.068 & \textbf{0.609} $\pm$ 0.054 \\
    \bottomrule
  \end{tabular}
\end{table}

\subsubsection{Multi-modal Fusion}

The multimodal fusion results are presented in Table~\ref{tab:multimodal_metrics}.
The naive concatenation approach performed the best with 79.2\% AUC-ROC, followed by Cross-time and Cross-modal Attention (ViT) approach with 73.5\% AUC-ROC.
We selected the naive feature concatenation approach for further downstream analysis.

\begin{table}[t]
  \caption{Performance for multimodal fusion in detecting challenging behaviors. Best and second-best scores per metric are \textbf{bolded} and \underline{underlined}, respectively.}
  \label{tab:multimodal_metrics}
  \centering
  \begin{tabular}{lcccc}
    \toprule
    \textbf{Fusion Method} & \textbf{AUC-ROC} & \textbf{Precision} & \textbf{Recall} & \textbf{F1 Macro} \\
    \midrule
    Naive Concatenation & \textbf{0.793} $\pm$ 0.093 & \textbf{0.300} $\pm$ 0.138 & \textbf{0.576} $\pm$ 0.187 & \textbf{0.633} $\pm$ 0.081 \\
    Temporal Self-Attention (ViT) & 0.716 $\pm$ 0.097 & 0.266 $\pm$ 0.133 & \underline{0.480} $\pm$ 0.165 & 0.607 $\pm$ 0.070 \\
    Cross-time Cross-modal Attention (ViT) & \underline{0.735} $\pm$ 0.105 & \underline{0.283} $\pm$ 0.112 & 0.460 $\pm$ 0.165 & \underline{0.611} $\pm$ 0.067 \\
    \bottomrule
  \end{tabular}
\end{table}

\subsection{Predicting Challenging Behaviors in Profound Autism}
\begin{figure}[t]
  \centering
  \includegraphics[width=0.6\textwidth]{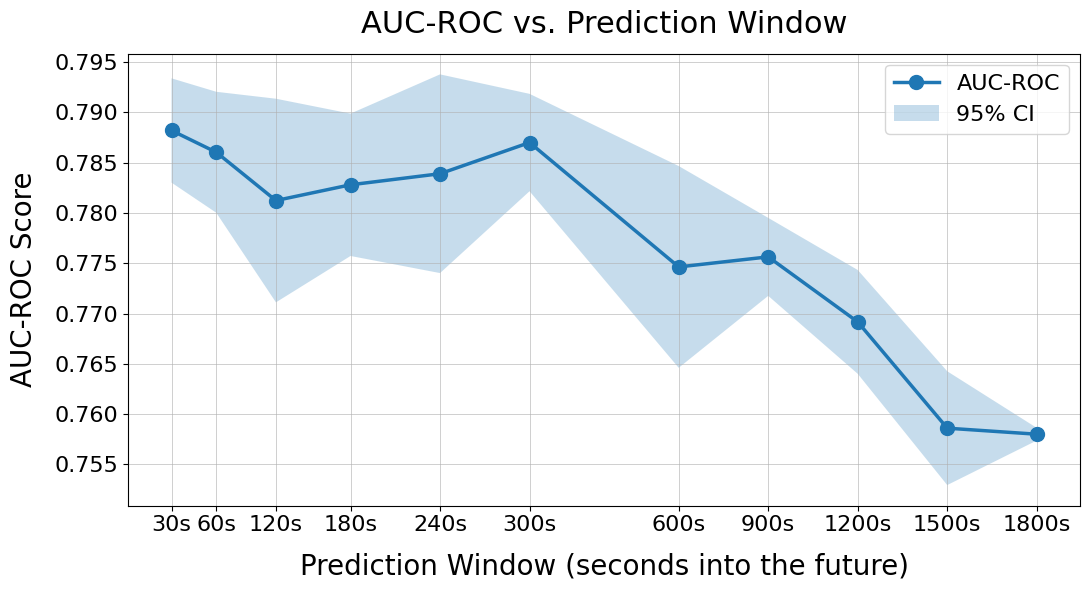}
  \caption{AUC-ROC scores with increasing prediction window for challenging behaviors.
  Performance decreases as the prediction horizon increases, indicating that predicting behaviors further into the future becomes progressively more challenging.
  }
  \label{fig:prediction_performance}
\end{figure}

\autoref{fig:prediction_performance} illustrates how the model’s predictive performance changes as the prediction horizon increases from 30 seconds to 1800 seconds (30 minutes).
Even at the longest horizon of 30 minutes, the AUC-ROC remained above 75\%.
The results at a 10-minute prediction horizon indicate that the multimodal fusion approach can maintain appropriate performance with an average AUC‑ROC of 0.78 with a 95\% CI of ±0.10, demonstrating reliable prediction of challenging behaviors. 
F1‑score, Precision, and recall were 0.39 ± 0.15, 0.31 ± 0.15, and 0.55 ± 0.14, respectively, reflecting a deliberate bias toward sensitivity in detecting challenging behaviors.


\subsubsection{Per-subject Performance for Prediction of Challenging Behavior}

\autoref{tab:subject_performance} summarizes 10-minute prediction performance for each subject.
Depending on the subject, the performance varied between 65\% and 90\% AUC-ROC.

\begin{table}[t]
  \caption{Per-subject performance metrics for prediction (10 mins)}
  \label{tab:subject_performance}
  \centering
  \begin{tabular}{lccccc}
    \toprule
    \textbf{Subject} & \textbf{Sensor location} & \textbf{AUC-ROC} & \textbf{Precision} & \textbf{Recall} & \textbf{F1 Macro} \\
    \midrule
    S01 & Right ankle & 0.765 & 0.203 & 0.588 & 0.570 \\
    S06 & Left ankle & 0.730 & 0.069 & 0.489 & 0.496 \\
    S07 & Left wrist & 0.870 & 0.272 & 0.500 & 0.656 \\
    S08 & Left ankle & 0.664 & 0.065 & 0.410 & 0.513 \\
    S09 & Left ankle & 0.902 & 0.491 & 0.770 & 0.748 \\
    S10 & Left ankle & 0.650 & 0.367 & 0.620 & 0.574 \\
    S11 & Left ankle & 0.651 & 0.395 & 0.220 & 0.594 \\
    S12 & Right ankle & 0.686 & 0.094 & 0.196 & 0.520 \\
    S13 & Left ankle & 0.758 & 0.357 & 0.073 & 0.496 \\
    \bottomrule
  \end{tabular}
\end{table}

\subsubsection{Feature Importance Analysis and Explainability for Prediction of Challenging Behavior}

\begin{figure}[t]
  \centering
  \includegraphics[width=0.6\textwidth]{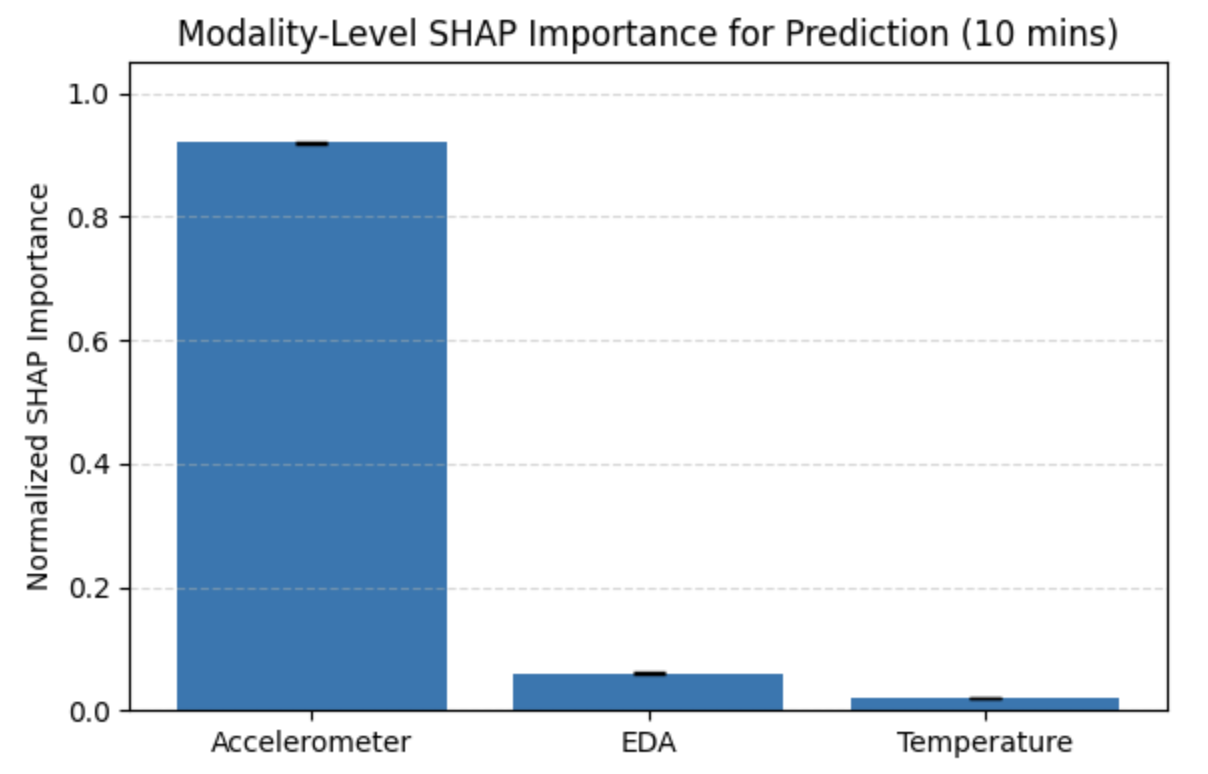}
  \caption{MM-SHAP-based modality importance for the multimodal model. Bars indicate the normalized mean contribution of accelerometer, EDA, and skin temperature signals to challenging behavior detection across cross-validation folds.}
  \label{fig:shap}
\end{figure}

\begin{figure}[t]
\centering
\subcaptionbox{Hands on ears behavior}{%
    \includegraphics[width=0.45\textwidth]{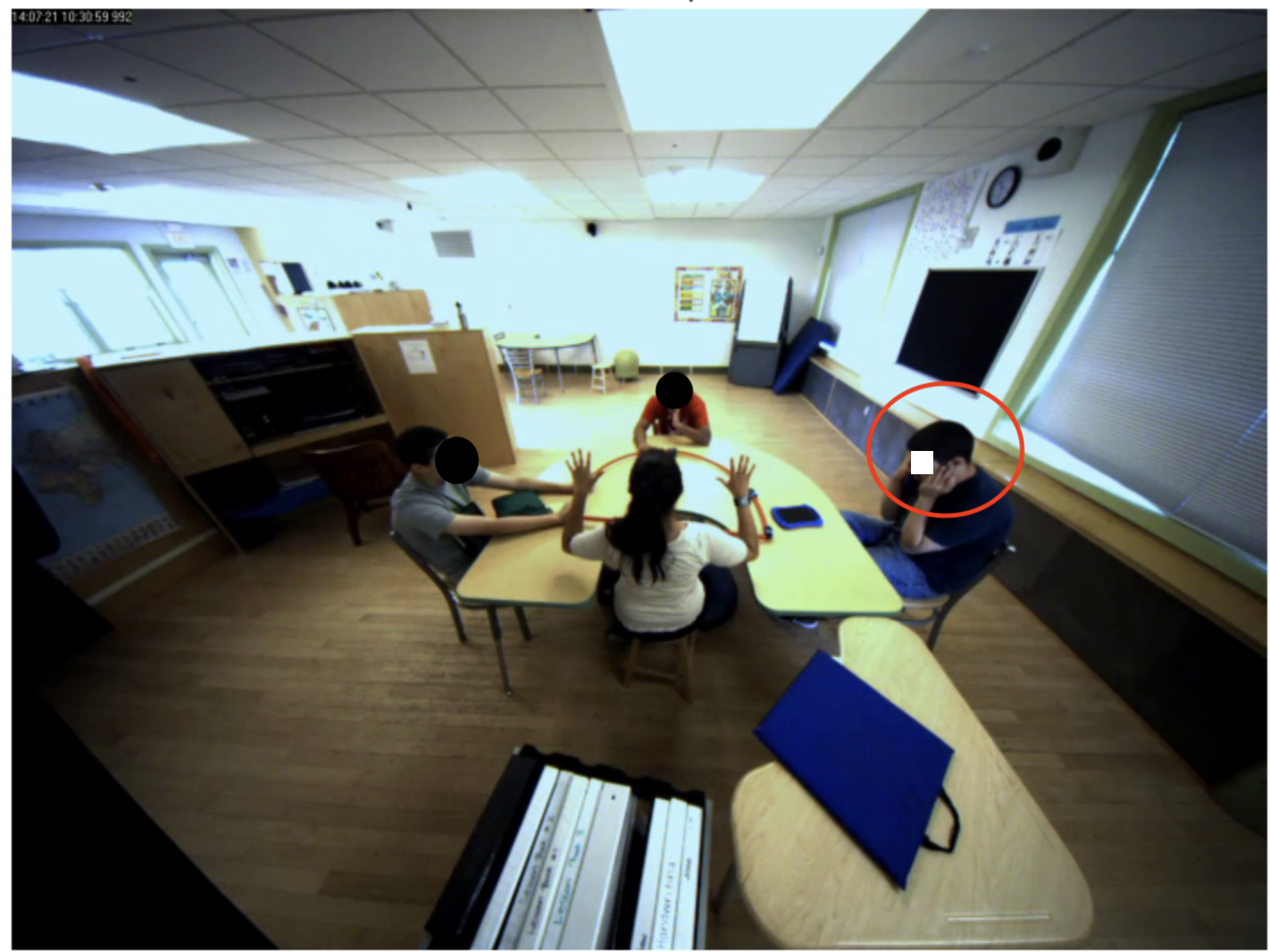}}
\subcaptionbox{Head tapping behavior}{%
    \includegraphics[width=0.45\textwidth]{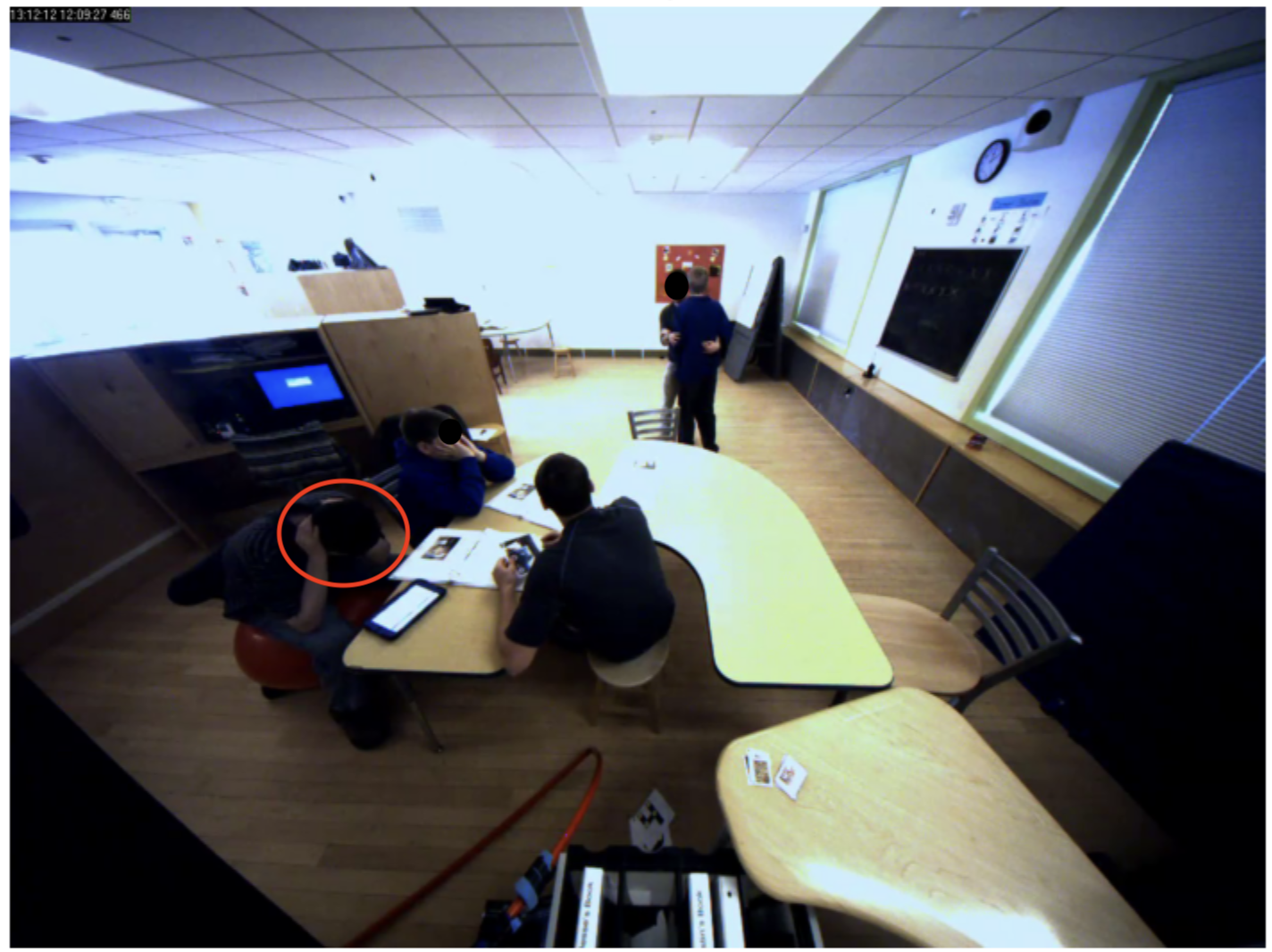}}
\subcaptionbox{Grad-CAM: Hands on ears}{%
    \includegraphics[width=0.45\textwidth]{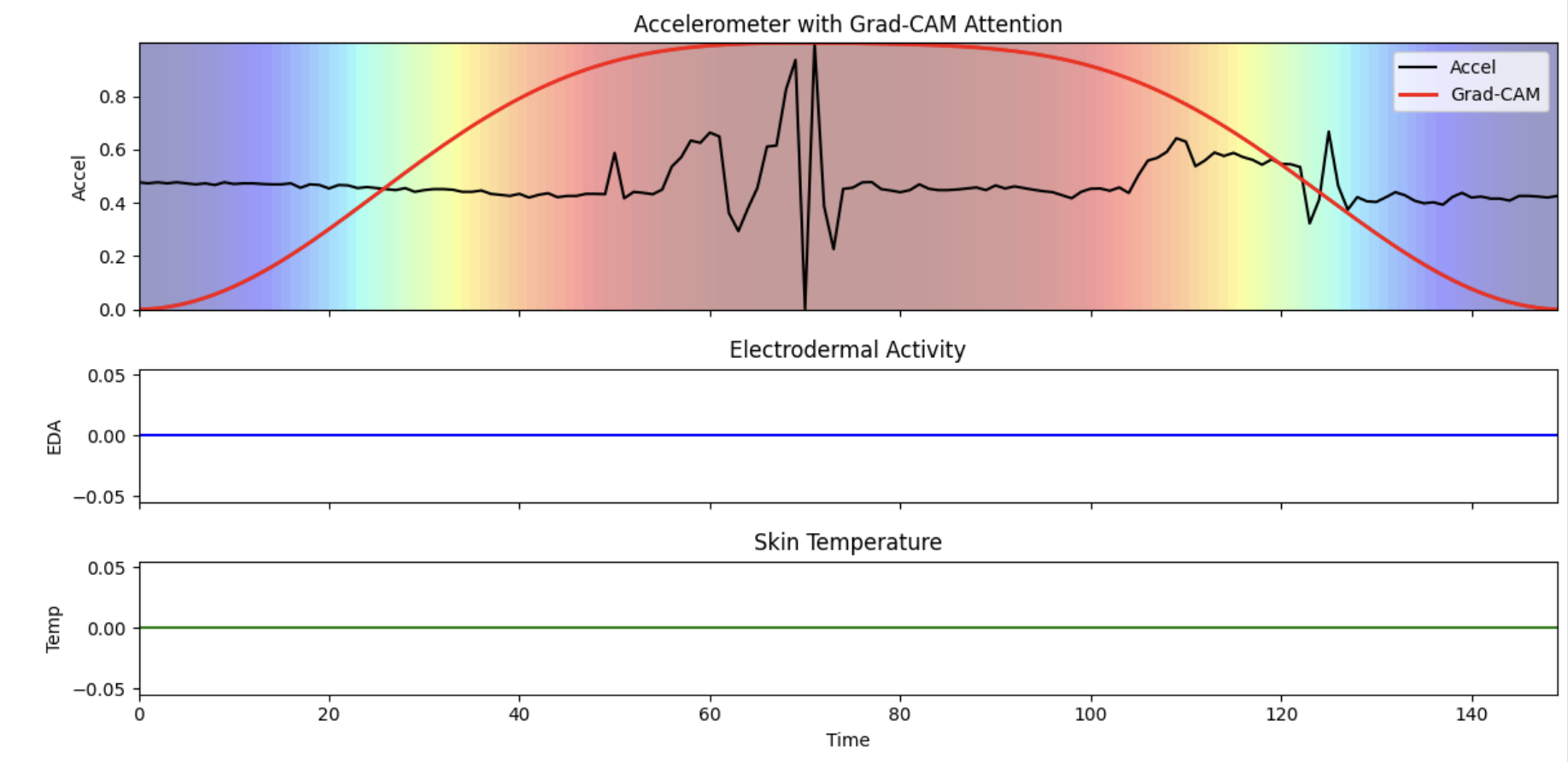}}
\subcaptionbox{Grad-CAM: Head tapping}{%
    \includegraphics[width=0.45\textwidth]{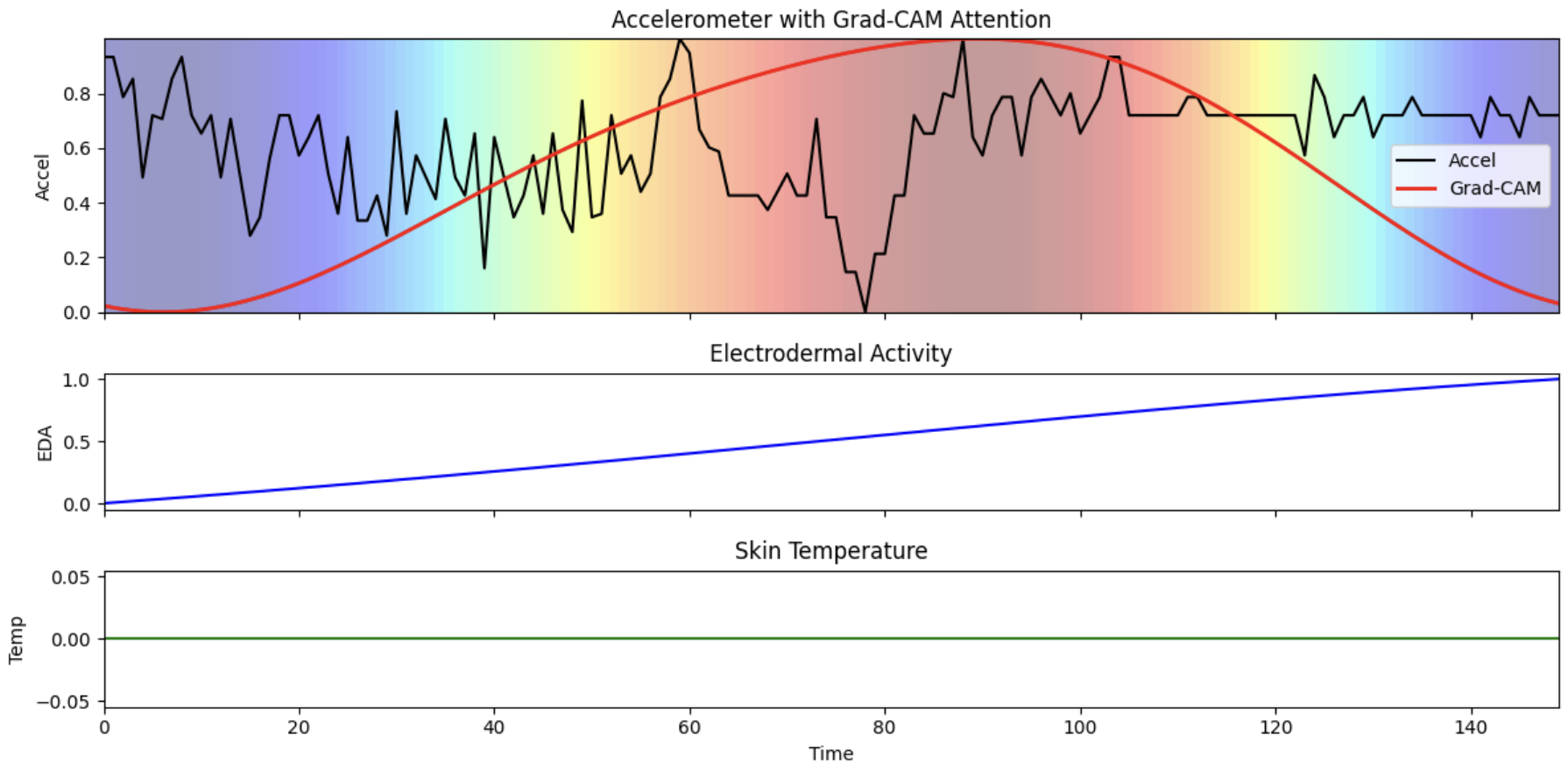}}
\caption{Grad-CAM visualizations for two representative behavior samples: (a,c) ``Hands on ears'' (Aggression) and (b,d) ``Head tapping'' (Self-Injurious Behavior). Rows (a,b) show the raw accelerometer trace; rows (c,d) overlay the Grad-CAM activation heatmap (red = high attention), highlighting the temporal segments most influential to the model's prediction. Peaks in attention align with periods of elevated motion intensity preceding or during the annotated behavior.}
\label{fig:gradcam_combined}
\end{figure}

\autoref{fig:shap} summarizes the average importance scores, revealing that accelerometer signals dominate model decisions, while EDA and temperature provide only marginal contributions.
\autoref{fig:gradcam_combined} shows that the Grad-CAM visualizations (\textcolor{red}{RED}) highlight temporal regions with high activation corresponding to annotated behaviors, along with accelerometer readings (\textcolor{blue}{BLUE}).

\subsubsection{Multiclass Prediction for Challenging Behaviors} 

\begin{figure}[t]
  \centering
  \subcaptionbox{Four-class multiclass classification (None, Aggression, SIB, Stereotypy)}{
    \includegraphics[width=0.45\textwidth]{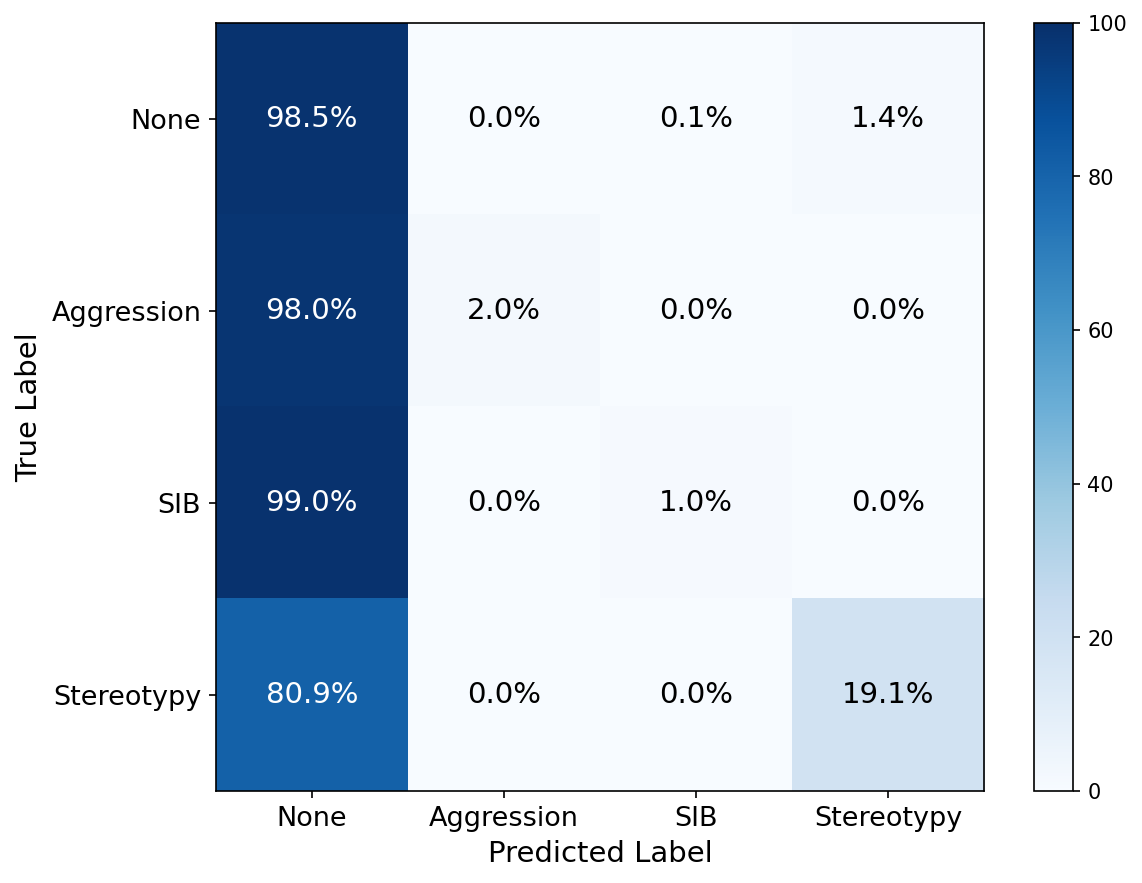}
  }
  \hfill
  \subcaptionbox{Three-class classification (None, High-risk, Low-risk)}{
    \includegraphics[width=0.45\textwidth]{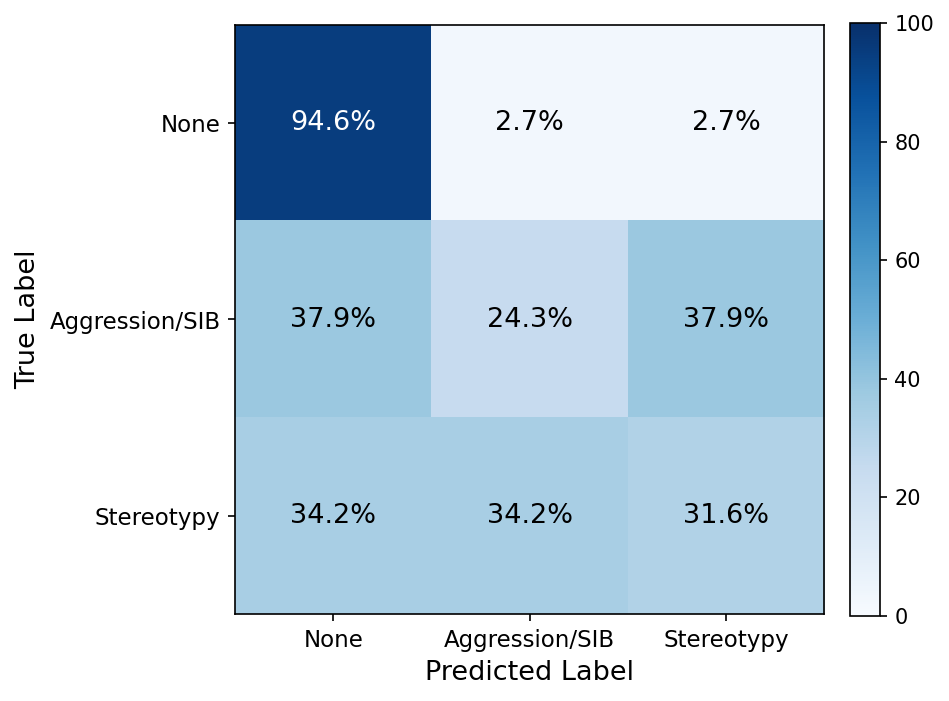}
  }
  \caption{Normalized confusion matrices for multiclass behavior prediction using the multimodal model. (a) Four-class classification reveals substantial confusion among minority behavior classes. (b) Grouping aggression and self-injurious behavior into a single high-risk class.}
  \label{fig:confusion_matrices}
\end{figure}

The four-class 10-minute prediction task (None, Aggression, SIB, Stereotypy) achieved an AUC-ROC of 0.65 using four-fold cross-validation.
The normalized confusion matrices for both settings are shown in \autoref{fig:confusion_matrices}.
In the four-class setting (\autoref{fig:confusion_matrices}a), the model achieves high specificity for the None class but exhibits diminished sensitivity for minority behavior classes, particularly Aggression, which is the rarest category in the dataset.
The coarser three-class formulation (None, High-risk, Low-risk) yielded an AUC-ROC of 0.53, which is above chance for a three-class problem, yet not sufficient to be useful, confirming that the severe imbalance between behavior categories fundamentally limits fine-grained behavior type discrimination at this data scale.

\section{Discussion}

\subsection{Detecting Challenging Behaviors in Profound Autism}

\subsubsection{Per-modality Detection}

The unimodal experiments yielded important insights into the relative contributions of different physiological and motion-based signals.
As shown in Table~\ref{tab:per_modality_metrics}, accelerometer-based models exhibited the strongest discriminative capacity, with the HarNet model, pretrained with self-supervised learning (contrastive learning), achieving the highest AUC-ROC of 0.778.
The DeepConvLSTM and the pretrained transformer classifier yielded comparatively lower performance.
This suggests that while temporal modeling offers benefits, these models, pretrained with supervised learning, may struggle under severe class imbalance, which likely limits their ability to generalize to rare positive events.
This limitation is consistent with prior findings in comparison between supervised and self-supervised  learning in wearable sensor activity recognition, where imbalanced datasets have been shown to degrade deep model performance \cite{Qureshi2025imbalanced,Lim2021Resampling}.

Electrodermal activity models, particularly the convolutional autoencoder, achieved strong results with an F1 score of 0.604 and the highest recall of 0.861, indicating high sensitivity to behavioral cues.
However, the lower AUC-ROC (0.62) and precision suggest a higher false-positive rate, possibly due to the non-specific nature of EDA responses and susceptibility to external confounds such as temperature, cognitive load, or emotional changes unrelated to challenging behaviors.
EDA reflects autonomic nervous system arousal, and prior studies have noted that while EDA is sensitive to physiological changes, it may lack specificity because multiple affective states can elevate skin conductance \cite{melander_measuring_2017}.
The handcrafted EDA features performed poorly across metrics, highlighting the limitations of shallow statistical summaries in capturing subtle phasic dynamics over time.
Skin temperature, although less dynamic, showed competitive results when modeled with the CNN-based architecture.
It achieved an F1 score of 0.609 and a precision of 0.588, achieving statistically similar performance to EDA in precision and overall F1 score.
Handcrafted temperature features underperformed, reinforcing the importance of learned representations for physiological signal modeling, consistent with findings in affective computing and stress recognition \cite{damelio_emotion_recognition_systems_2025}.
The performance differences across modalities likely reflect inherent signal characteristics and annotation alignment.
Motion signals directly capture behavioral escalation through increased physical activity and are typically high-frequency and unambiguous, whereas EDA and temperature are indirect and slower autonomic correlates that are more variable and susceptible to non-behavioral fluctuations.
This pattern is reflected in the higher recall but lower precision of EDA-based models and the moderate, yet balanced, performance of temperature-based models.
These findings motivated the development of multimodal fusion approaches to combine complementary motion and autonomic signals for improved behavior detection.

\subsubsection{Multi-modal detection}

The fusion of acceleration, electrodermal activity, and skin temperature through naive feature concatenation led to the highest overall performance, achieving an AUC-ROC of $0.792$ and the highest F1 score ($0.633$) among all multimodal variants (see Table~\ref{tab:multimodal_metrics}).
This improvement over unimodal models indicates that physiological and movement signals provide complementary cues for behavior prediction, with accelerometry capturing overt motor escalation and EDA and temperature contributing supportive autonomic context.
Prior studies in affective and health sensing have shown that integrating motion and physiological signals enhances detection performance in stress and arousal classification tasks \cite{le_tran_thuan_machine_2024,logacjov_machine_2024}, supporting the utility of multimodal fusion in challenging behavioral modeling.
The Transformer-based multimodal fusion approaches, including temporal self-attention and cross-time cross-modal attention, demonstrated consistently lower performance compared to naive feature concatenation.
We attribute this to two interacting factors.
First, the limited size and highly imbalanced nature of our dataset constrains the ability of self-attention layers to learn stable cross-modal relationships.
With only ${\sim}$110.7 hours of data from 9 participants and severe class imbalance, the transformer heads lack sufficient positive-class diversity to learn meaningful attention patterns, and risk overfitting to spurious cross-modal correlations within the small training dataset.
Second, attention-based architectures such as Vision Transformers have demonstrated strong performance when trained on large and balanced datasets but are known to be sensitive to data scarcity and noise, often requiring extensive regularization or large-scale pretraining to prevent overfitting \cite{dosovitskiy2021imageworth16x16words}.

\subsection{Predicting Challenging Behaviors in Profound Autism}

\subsubsection{Impact of Prediction Horizon}

\autoref{fig:prediction_performance} showed a gradual decline in AUC-ROC as the prediction window increased from 30 seconds to 30 minutes.
This trend is expected, as predicting behaviors farther into the future introduces more uncertainty due to dynamic environmental and contextual factors \cite{oreshkin_n-beats_2020}.
Short-term predictions leverage immediate physiological and motion precursors to behaviors, while long-term predictions require modeling delayed and nonlinear dependencies that are harder to capture with limited data.
Empirically, the 10-minute horizon represented the longest interval at which prediction performance remained comparable to detection metrics before a marked degradation in AUC-ROC and F1 score was observed \cite{goodwin_wearable_2023}.
This result may underscore the feasibility of behavioral predictions, but also calls for further study with larger, temporally diverse datasets to potentially improve generalization to longer prediction horizons.

\subsubsection{Clinical Implications of Precision - Recall Trade-offs}

The binary prediction model at a 10-minute horizon achieved a recall of 0.55 but a precision of only 0.31, implying that approximately two out of three alerts would be false positives.
Whether this trade-off is acceptable depends on the clinical context and the cost asymmetry between missed events and false alarms.
In special education classrooms, failing to anticipate a self-injurious or aggressive episode carries a high cost of potential physical harm to the student, peers, and staff whereas a false alert imposes a comparatively lower cost, such as unnecessary redirection of teacher attention.
From this perspective, a sensitivity-biased operating point is potentially clinically defensible, as it prioritizes safety over alert specificity.
However, if false-alarm rates are persistently high, teachers may develop ``alert fatigue,'' progressively ignoring notifications and undermining the system's utility \cite{ancker_effects_2017_alert_fatigue}.

\subsubsection{Per-subject Performance on Behavior Prediction Task}

We analyzed subject-wise prediction performance for a 10-minute horizon to evaluate consistency and potential disparities across individuals.
As shown in Table~\ref{tab:subject_performance}, model AUC-ROC scores ranged from 0.650 (S10) to 0.902 (S09), with no clear trend based on age.
Notably, S07 (wrist placement) and S09 (left ankle placement) both achieved strong AUC-ROC values of 0.870 and 0.902, respectively, indicating that placement site alone did not determine performance, owing to the adaptation of a foundation model pretrained on a large wearable dataset.
Prediction accuracy appeared to correlate more strongly with behavior prevalence and label density, as shown in Figure~\ref{fig:subject_duration}.
Subjects with higher behavioral event frequency, such as S09 (F1 score: 0.748), demonstrated stronger performance.
S10 (F1 score: 0.574), though moderate in absolute terms, outperformed subjects with sparser labels, consistent with positive-event density supporting reliable temporal learning.
Lower F1 scores were observed for S06 (F1 score: 0.496) and S13 (F1 score: 0.496), who had fewer annotated behavioral events and lower inter-session annotation consistency, highlighting the impact of label sparsity on model reliability.

\subsubsection{Feature Importance Analysis}

As shown in \autoref{fig:shap}, MM-SHAP \cite{Parcalabescu_2023} importance scores aggregated across test folds indicate that accelerometry dominated the prediction process, contributing an average of 0.90 to the final decision.
Electrodermal activity and skin temperature contributed less, 0.08 and 0.02, respectively, suggesting that motion signals remain the primary driver of early behavior prediction.
EDA and skin temperature reflect indirect autonomic responses that evolve more slowly and are influenced by multiple contextual factors, reducing their specificity to discrete behavioral events\cite{shin2025skintemp}.
The non-zero contribution of EDA implies that anticipatory autonomic changes may still provide predictive cues leading up to behavioral episodes.
\autoref{fig:gradcam_combined} visualizes temporal attention over input sequences preceding behaviors such as ``hands on ears'' and ``head tapping''.
In both cases, the highest Grad-CAM\cite{Selvaraju_2019} activations align with pronounced accelerometer fluctuations, particularly around sharp motion transitions and rhythmic acceleration bursts in the head tapping example.
The corresponding EDA and skin temperature signals exhibit either flat or slowly varying trends during these intervals, indicating that the attention maps correlate more strongly with movement dynamics than with autonomic channels.

\subsubsection{Multiclass Prediction for Challenging Behaviors}

\autoref{fig:confusion_matrices}a presents the normalized confusion matrix for the four-class prediction task (None, Aggression, SIB, Stereotypy).
While the model achieves an AUC-ROC of 0.65, this discriminative capacity is driven almost entirely by the dominant None class, which is correctly classified at 98.5\%.
The model fails to detect Aggression  (2\% true positive rate) and identifies only 1.0\% of SIB instances, with the remaining 99.0\% misclassified as None.
Stereotypy fares marginally better at 19.1\% recall, though the majority of instances (80.9\%) are still absorbed by the None class.
These results indicate that the four-class model effectively collapses to a binary  behavior vs.\ no behavior  detector, unable to learn type-specific precursors for rare behavior categories.
Consolidating Aggression and SIB into a single high-risk category (\autoref{fig:confusion_matrices}b) does not resolve this limitation.
The resulting three-class model yields an AUC-ROC of 0.53, which is above chance for a three-class problem, yet not sufficiently useful, confirming that behavior-type discrimination is not feasible under the current data regime.
Classification of the None class remains robust at 94.6\%, but both the high-risk (24.3\% recall) and Stereotypy (31.6\% recall) classes exhibit near-uniform confusion across all three categories, consistent with random assignment among minority classes.
These multiclass results constitute a finding that establishes a clear boundary condition for this line of work.

The failure is attributable to the extreme label imbalance inherent in the dataset: Aggression accounts for fewer than 5\% of all behavior events (\autoref{fig:behavior_duration}), providing insufficient examples for the model to learn class-discriminative temporal patterns.
This is consistent with prior findings that severe class imbalance degrades deep model performance in wearable activity recognition, even when resampling strategies are employed \cite{Qureshi2025imbalanced,Lim2021Resampling}.
In contrast to the binary prediction task, where collapsing all behaviors into a single positive class yields sufficient training signal (AUC-ROC of 0.78 at 10 minutes), the multiclass formulation distributes an already sparse positive set across multiple categories, compounding the data scarcity problem.
Future work should prioritize expanding data collection for underrepresented behavior types particularly aggression and SIB across a larger participant cohort, and investigate class conditional augmentation or few-shot learning strategies to enable fine-grained behavior-type prediction \cite{zwilling2022prediction}.

\subsection{Limitations and Future Work}

While the current findings demonstrate the feasibility of multimodal wearable sensing for predicting challenging behaviors, several limitations warrant consideration.
First, the dataset is relatively small and highly imbalanced, with challenging behavior events, particularly aggression and SIB, occurring infrequently and unevenly across participants.
This class imbalance likely constrained the model's generalization capacity and introduced bias toward the majority (non-event) class.
As shown in Table~\ref{tab:subject_performance}, detection performance varied substantially across individuals, with precision ranging from 0.065 to 0.491 and recall from 0.073 to 0.770.
These disparities appear to be driven primarily by the quantity and diversity of behavior episodes per participant rather than by demographic characteristics such as age.
Ongoing work aims to expand data collection across a larger cohort and multiple sessions in diverse settings across TCFD sites, with the goal of improving model robustness and reducing systematic bias.
Second, while short-horizon prediction yielded strong discriminative performance (AUC-ROC > 0.78), accuracy degraded progressively at longer prediction horizons (e.g., 30 minutes), as shown in \autoref{fig:prediction_performance}.
This reflects a fundamental challenge in behavioral prediction: as the prediction horizon extends, the likelihood of disruption by unmodeled contextual dynamics increases~\cite{zwilling2022prediction}.
To address this limitation, future work will explore multimodal fusion incorporating video, skeletal pose estimation, and wearable signals, providing richer behavioral context and enabling the capture of fine-grained movement patterns that may serve as early precursors to challenging behavior.
Integration of auxiliary contextual signals—such as time-of-day, structured activity schedules, and ambient environmental cues—will also be investigated to better support long-range behavioral prediction.
Third, generalization to external cohorts presents a challenging cross-cohort evaluation scenario.
We preliminarily studied the trained model on an external dataset that differs from our training data in multiple dimensions simultaneously: sensor device (Affectiva Q-Sensor vs. Empatica E4), sampling rates (uniform 30 Hz vs. 32/4/4 Hz for accelerometry, EDA, and temperature), sensor placement (predominantly ankle vs. exclusively wrist).
Due to these differences, the model did not generalize to the external dataset, and future work will include adapting and fine tuning to datasets using data centric and model centric methods for cross-domain generalization \cite{cai2025generalizablehumanactivityrecognition}.

\section{Conclusion}

Proactive intervention for challenging behaviors in children with profound autism demands tools that can anticipate, not merely document, challenging behavior episodes in the complex, dynamic environments where these individuals spend their days.
This work demonstrates that such anticipation is feasible in a real-world special education classroom: by fine-tuning pretrained multimodal representation models on accelerometry, electrodermal activity, and skin temperature collected from nine participants, our system achieved an AUC-ROC of 0.78 at an educationally useful 10-minute prediction horizon for intervention, establishing a concrete operational window for teacher alerting.
Accelerometry emerged as the dominant predictive signal, as confirmed by MM-SHAP importance scores and Grad-CAM temporal activations that consistently highlighted motion-rich intervals preceding behaviors such as hand-to-ear gestures and head tapping, while EDA and skin temperature contributed complementary autonomic context that improved overall fusion performance.
Per-subject analysis further revealed that prediction accuracy correlates primarily with behavioral event density rather than age or sensor placement, with S07 (wrist-mounted) achieving AUC-ROC of 0.870, demonstrating that strong performance is attainable across placement sites when sufficient labeled data are available.
Collectively, these findings establish multimodal wearable sensing as a viable foundation for proactive behavior prediction in profound autism, with clear pathways toward real-time alert systems, and personalized fine-tuning pipelines.
Future integration of contextual signals such as video, skeleton poses, activity schedules, and environmental audio could extend reliable prediction horizons and ultimately improve safety and educational outcomes for one of the most underserved populations in special education.

\section*{Acknowledgments} 

The authors gratefully acknowledge the participants and their families for their consent and ongoing engagement throughout the study.
We thank the behavior analysts, research assistants, classroom teachers, and direct-care staff at TCFD for their dedication to data collection, video annotation, and day-to-day operational support, without which this work would not have been possible.
We also thank members of TCFD for helpful discussions on multimodal time-series modeling and clinical interpretation of challenging behaviors.
Any opinions, findings, and conclusions expressed in this article are those of the authors and do not necessarily reflect the views of the funding agencies or participating institutions.
Hyeokhyen Kwon and Gari D. Clifford are partially funded by the National Institute on Deafness and Other Communication Disorders (grant No. 1R21DC021029-01A1).
Hyeokhyen Kwon is partially funded by Georgia CTSA Pilot Grants Program.

\bibliographystyle{ACM-Reference-Format}
\bibliography{Wearables}

\end{document}